\documentclass[10pt,twocolumn,letterpaper]{article}

\usepackage{latex/wacv}
\usepackage{times}
\usepackage{epsfig}
\usepackage{graphicx}
\usepackage{amsmath}
\usepackage{amssymb}
\usepackage{subcaption}
\usepackage{url}
\usepackage{import} 
\usepackage{enumitem}

\usepackage{pgfplots}
\pgfplotsset{compat=1.9}

\wacvfinalcopy 

\def\wacvPaperID{53} 

\ifwacvfinal\fi
\begin{document}

\title{UG$^2$: a Video Benchmark for Assessing the Impact of Image Restoration and Enhancement on  Automatic Visual Recognition \vspace{-5mm} \thanks{* denotes equal contribution}}

\author{\vspace{-1.8mm}\parbox{16cm}{\centering
    {\large Rosaura G. Vidal*$^1$, 
    Sreya Banerjee*$^1$, Klemen Grm\textsuperscript{2}, Vitomir \v{S}truc\textsuperscript{2} and Walter J. Scheirer\textsuperscript{1}}\\
    {\normalsize
    $^1$ Dept. of Computer Science \& Engineering, University of Notre Dame, USA\\
    $^2$ University of Ljubljana, Slovenia\\
        \tt\small \{rvidalma, sbanerj2, walter.scheirer\}@nd.edu\vspace{-0.5mm}\\
        \tt\small \{klemen.gr, vitomir.struc\}@fe.uni-lj.si
    }   
    }
    }
\renewcommand\footnotemark{}
\renewcommand\footnoterule{}

\maketitle
\ifwacvfinal\thispagestyle{empty}\fi

\begin{abstract}\vspace{-1.8mm}
Advances in image restoration and enhancement techniques have led to discussion about how such algorithms can be applied as a pre-processing step to improve automatic visual recognition. In principle, techniques like deblurring and super-resolution should yield improvements by de-emphasizing noise and increasing signal in an input image. But the historically divergent goals of computational photography and visual recognition communities have created a significant need for more work in this direction. To facilitate new research, we introduce a new benchmark dataset called UG$^2$, which contains three difficult real-world scenarios: uncontrolled videos taken by UAVs and manned gliders, as well as controlled videos taken on the ground. Over 150,000 annotated frames for hundreds of ImageNet classes are available, which are used for baseline experiments that  assess the impact of known and unknown image artifacts and other conditions on common deep learning-based object classification approaches. Further, current image restoration and enhancement techniques are  evaluated by determining whether or not they improve baseline classification performance. Results show that there is plenty of room for algorithmic innovation, making this dataset a useful tool going forward. \vspace{-3mm} 


\end{abstract}

\section{\vspace{-1.8mm}Introduction} \label{Introduction}
\begin{figure}[t]
\begin{center}
\includegraphics[width=0.4\textwidth]{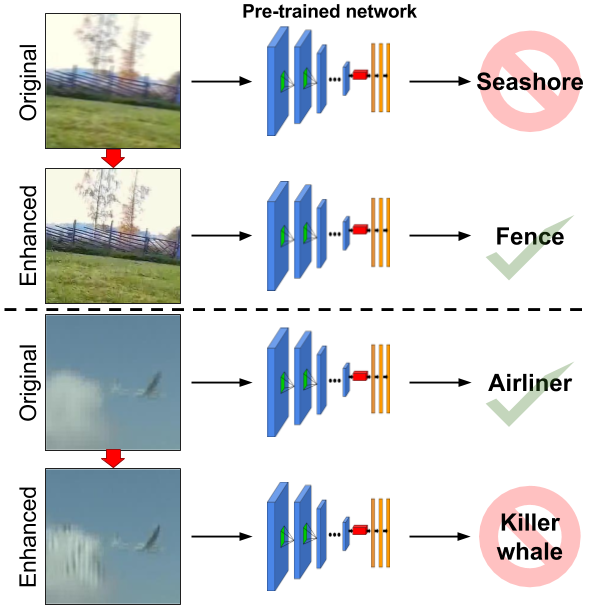}
\end{center}\vspace{-5mm}
   \caption{(Top) In principle, image restoration and enhancement techniques should improve visual recognition performance by creating higher quality inputs for recognition models. This is the case when a Super Resolution Convolutional Neural Network~\cite{Chao:2014:SRCNN} is applied to the image in this panel. (Bottom) In practice, we often see the opposite effect --- especially when new artifacts are unintentionally introduced, as in this  application of Deep Deblurring~\cite{Su:2016:DBN}. We describe a new video dataset (Sec.~\ref{Dataset}) for the study of problems with algorithm and data interplay (Sec.~\ref{Results}) like this one.}
\label{fig:Teaser}
\vspace{-5mm}
\end{figure}


To build a visual recognition system for any application these days, one's first inclination is to turn to the most recent machine learning breakthrough from the area of deep learning, which no doubt has been enabled by access to millions of training images from the Internet. But there are many circumstances where such an approach cannot be used as an off-the-shelf component to assemble the system we desire, because even the largest training dataset does not take into account all of the artifacts that can be experienced in the environment. As computer vision pushes further into real-world applications, what should a software system that can interpret images from sensors placed in any unrestricted setting actually look like? 

First, it must incorporate a set of algorithms, drawn from the areas of computational photography and machine learning, into a processing pipeline that corrects and subsequently classifies images across time and space. Image restoration and enhancement algorithms that remove corruptions like blur, noise, and mis-focus, or manipulate images to gain resolution, change perspective, and compensate for lens distortion are now commonplace in photo editing tools. Such operations are necessary to improve the quality of raw images that are otherwise unacceptable for recognition purposes. But they must be compatible with the recognition process itself, and not adversely affect feature extraction or classification (Fig.~\ref{fig:Teaser}). 

Remarkably, little thought has been given to image restoration and enhancement algorithms for visual recognition --- the goal of computational photography thus far has simply been to make images look appealing after correction~\cite{zeyde2010single,bevilacqua2012low,timofte2014,huang2015single,Su:2016:DBN}. It remains unknown what impact many transformations have on visual recognition algorithms. To begin to answer that question, exploratory work is needed to find out which image pre-processing algorithms, in combination
with the strongest features and supervised machine learning approaches, are promising candidates for different problem domains.

One popular problem that contains imaging artifacts typically not found in computer vision datasets crawled from the web is the interpretation of images taken by aerial vehicles~\cite{tomic2012toward,reilly2013shadow}. This task is a key component of a number of applications including robotic navigation, scene reconstruction, scientific study, entertainment, and visual surveillance. 
Images captured from aerial vehicles tend to present a wide variety of artifacts and optical aberrations. These can be the product of weather and scene conditions (\eg, rain, light refraction, smoke, flare, glare, occlusion), movement (\eg, motion blur), or the recording equipment (\eg, video compression, sensor noise, lens distortion, mis-focus). 
How do we begin to address the need for image restoration and enhancement algorithms that are compatible for visual recognition in a scenario like this one?

In this paper, we propose the use of a benchmark dataset captured in realistic settings where image artifacts are common, as opposed to more typical imagery crawled from social media and web-based photo albums. Given the current popularity of aerial vehicles within computer vision, we suggest the use of data captured by UAVs and manned gliders as a relevant and timely challenge problem. But this can't be the only data we consider, as we do not have knowledge of certain scene parameters that generated the artifacts of interest. Thus we suggest that ground-based video with controlled scene parameters and target objects is also essential. And finally, we need a set of protocols for evaluation that move away from a singular focus on perceived image quality to include classification performance. By combining all of these elements, we have created a new dataset dubbed UG$^2$ (UAV, Glider, and Ground), which consists of hundreds of videos and over 150,000 annotated frames spanning hundreds of ImageNet classes~\cite{ILSVRC15}. 



In summary, the contributions of this paper are: \vspace{-3mm} 
\begin{itemize}[noitemsep]
\item A new video benchmark dataset representing both ideal conditions and common aerial image artifacts, which we make available to facilitate new research and to simplify the reproducibility of experimentation\footnote{See the Supplemental Material for example videos from the dataset.The dataset can be accessed at: \url{https://goo.gl/AjA6En}.}. 
\item An extensive evaluation of the influence of image aberrations and other problematic conditions on common object recognition models including VGG16 and VGG19~\cite{VGG:2014}, InceptionV3~\cite{Inception:2015}, and ResNet50~\cite{ResNet50:2015}. 
\item An analysis of the impact and suitability of basic and state-of-the-art image and video processing algorithms used in conjunction with common object recognition models. In this work, we look at  deblurring~\cite{Su:2016:DBN, Nah:2016:DSDD}, image interpolation, and super-resolution~\cite{Chao:2014:SRCNN, Kim:2016:VDSR}.   
\end{itemize} \vspace{-4mm} 

\section{\vspace{-1.8mm}Related work} \label{Related-work}


\textbf{Datasets.} The areas of image restoration and enhancement have a long history in computational photography, with associated benchmark datasets that are mainly used for the qualitative evaluation of image appearance. These include very small test image sets such as Set5~\cite{bevilacqua2012low} and Set14~\cite{zeyde2010single}, and the set of blurred images introduced by Levin \etal~\cite{levin2009understanding}. Datasets containing more diverse scene content have been proposed including Urban100~\cite{huang2015single} for enhancement comparisons and LIVE1~\cite{sheikh2006statistical} for image quality assessment. While not originally designed for computational photography, the Berkeley Segmentation Dataset has been used by itself~\cite{huang2015single} and in combination with LIVE1~\cite{yang2014single} for enhancement work. The popularity of deep learning methods has increased demand for training and testing data, which Su \etal provide as video content for deblurring work~\cite{Su:2016:DBN}. Importantly, none of these datasets were designed to combine image restoration and enhancement with recognition for a unified benchmark.



Most similar to the dataset we introduce in this paper are various large-scale video surveillance datasets, especially those which provide a ``fixed" overhead view of urban scenes~\cite{CAVIAR:Dataset:2004, grgic2011scface, CUHK:Dataset:2014, TISI:Dataset:2013}. However, these datasets are primarily meant for other research areas (\eg, event/action understanding, video summarization,  face recognition) and are ill-suited for object recognition tasks, even if they share some common imaging artifacts that impair recognition as a whole. 
With respect to data collected by aerial vehicles, the VIRAT Video Dataset~\cite{Virat:Dataset:2011} contains ``realistic, natural and challenging (in terms of its resolution, background clutter, diversity in scenes)" imagery for event recognition. Other datasets including aerial imagery are the UCF Aerial Action Data Set~\cite{UCFAA}, UCF-ARG~\cite{UCFARG}, UAV123~\cite{mueller2016benchmark}, and the multi-purpose dataset introduced by Yao \etal~\cite{LHI:Dataset:2007}. As with the computational photography datasets, none of these sets have specific protocols for image restoration and enhancement coupled with object recognition. 


\begin{table}
\centering
\begin{center}
\begin{tabular}{|c|c|c|c|}
\hline
\textbf{Collection} & \textbf{UAV} & \textbf{Glider}  & \textbf{Ground}   \\
\hline\hline

Total Videos & 50 & 61 & 178 \\
Total Frames & 434,264 & 657,455 & 125,777 \\ 
Annotated videos & 30 & 30 & 136 \\
Extracted objects & 32,608 & 31,760 & 95,096 \\
ImageNet super-classes & 31 & 20 & 20 \\
Uncontrolled artifacts* & 12 & 9 & 3\\
Induced artifacts* & --- & --- & 13 \\

\hline
\end{tabular}
\end{center}
\vspace{-3mm}
\caption{Summary of the UG$^2$ dataset (See Supp. Tables~3 and 4 for a detailed breakdown of these conditions).}
\label{tab:dataset_summary}
\vspace{-6mm}
\end{table}

\textbf{Restoration and Enhancement for Recognition.} In this paper we consider the image restoration technique of deblurring, where the objective is to recover a sharp version $x'$ of a blurry image $y$ without knowledge of the blur parameters. When considering motion, the original sharp image $x$ is convolved with a blur kernel $k$: $y = k \ast x$~\cite{levin2011efficient}. Accordingly, the sharp image can be recovered through deconvolution~\cite{levin2007image, levin2009understanding, joshi2009image, levin2011natural} (see \cite{wang2014recent} for a comprehensive list of deconvolution techniques for deblurring) or methods that use multi-image aggregation and fusion~\cite{law2006lucky, matsushita2006full, cho2009fast}. Intuitively, if an image has been corrupted by blur, then deblurring should improve performance of recognizing objects in the image. An early attempt at unifying a high-level task like object recognition with a low-level task like deblurring was the Deconvolutional Network~\cite{zeiler2010deconvolutional, zeiler2011adaptive}. Additional work has been undertaken in face recognition~\cite{yao2008improving,nishiyama2009facial,zhang2011close}. In this work, we look at deep learning-based deblurring techniques~\cite{Nah:2016:DSDD, Su:2016:DBN} and a basic blind deconvolution method~\cite{kundur1996blind}.

With respect to enhancement, we focus on the specific technique of single image super-resolution, where an attempt is made at estimating a high-resolution image $x$ from a single low-resolution image $y$. The relationship between these images can be modeled as a linear transformation $y = Ax + n$, where $A$ is a matrix that encodes the processes of blurring and downsampling, and $n$ is a noise term~\cite{efrat2013accurate}. A number of super-resolution techniques exist including sparse representation~\cite{yang2008image, yang2010image}, Nearest Neighbor approaches~\cite{freeman2002example}, image priors~\cite{efrat2013accurate}, local self examples~\cite{freedman2011image}, neighborhood embedding~\cite{timofte2013anchored}, and deep learning \cite{Chao:2014:SRCNN,Kim:2016:VDSR}. Super-resolution can potentially help in object recognition by amplifying the signal of the target object to be recognized. Thus far, such a strategy has been limited to research in face recognition~\cite{lin2005face, lin2007super, hennings2008simultaneous, yu2011face, huang2011super, uiboupin2016facial, rasti2016convolutional} and person re-identification~\cite{jing2015super} algorithms for video surveillance data. Here we look at simple interpolation methods~\cite{Keys:1981:Interpolation} and deep learning-based super-resolution \cite{Chao:2014:SRCNN, Kim:2016:VDSR}.

\vspace{-3mm} 
\section{\vspace{-1.8mm}The UG$^2$ Dataset} \label{Dataset}

\begin{figure}[t]
    \centering
    \begin{subfigure}{0.40\textwidth}
        \centering
        \includegraphics[width=0.40\textwidth]{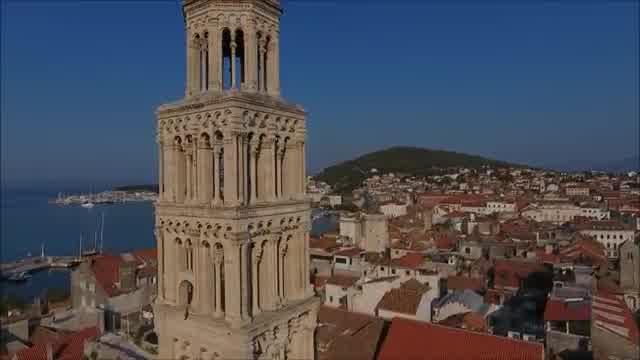}
        \includegraphics[width=0.40\textwidth]{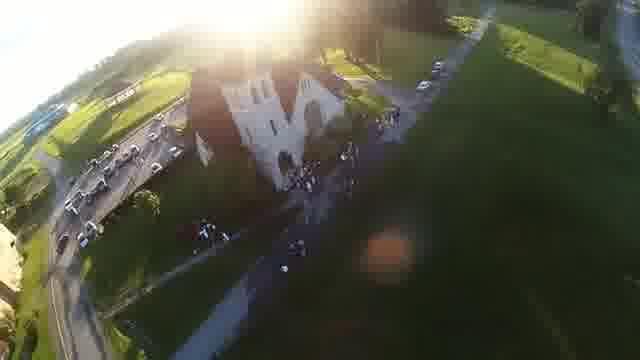}
        \caption{UAV Collection}
    \end{subfigure}\vskip 0mm
    \begin{subfigure}{0.40\textwidth}
         \centering
        \includegraphics[width=0.40\textwidth]{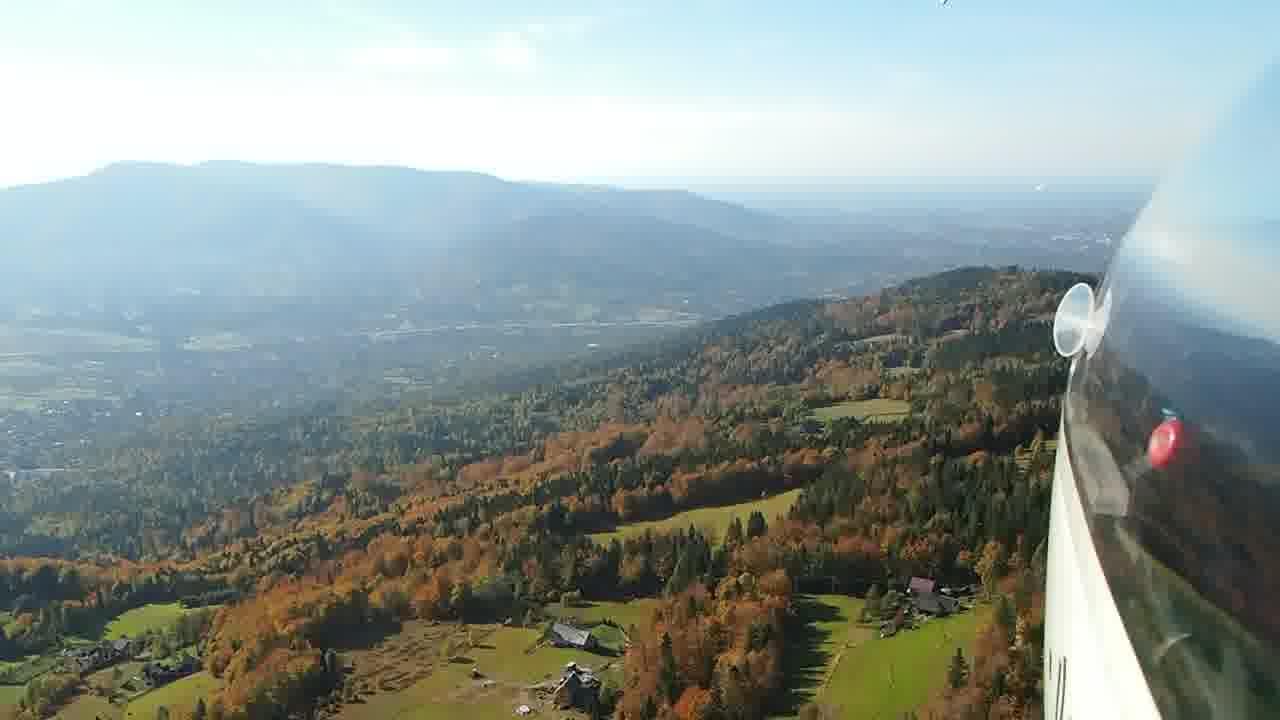}
        \includegraphics[width=0.40\textwidth]{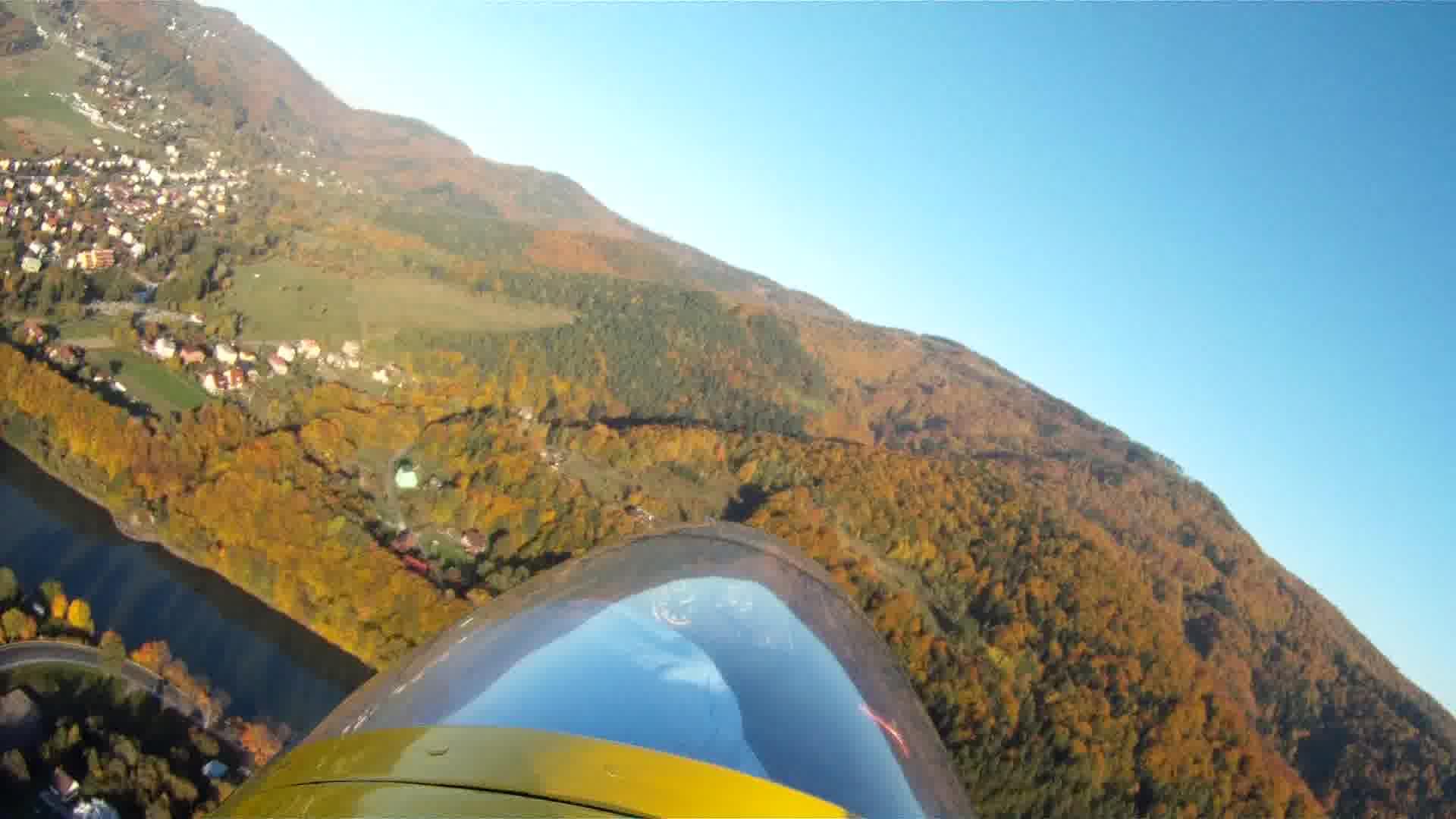}
        \caption{Glider Collection}
    \end{subfigure}\vskip 0mm
    \begin{subfigure}{0.40\textwidth}
        \centering
        \includegraphics[width=0.40\textwidth]{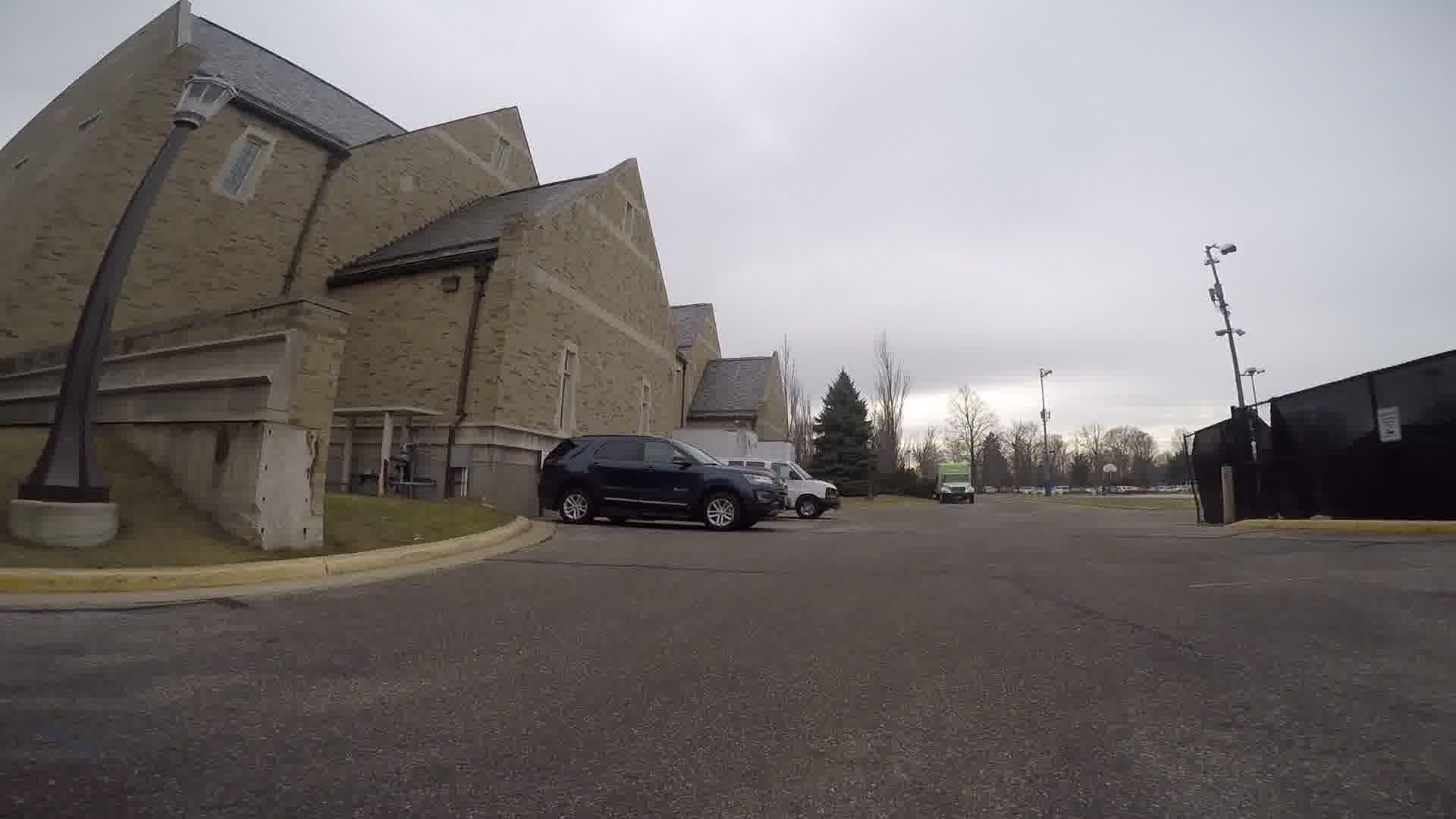}
        \includegraphics[width=0.40\textwidth]{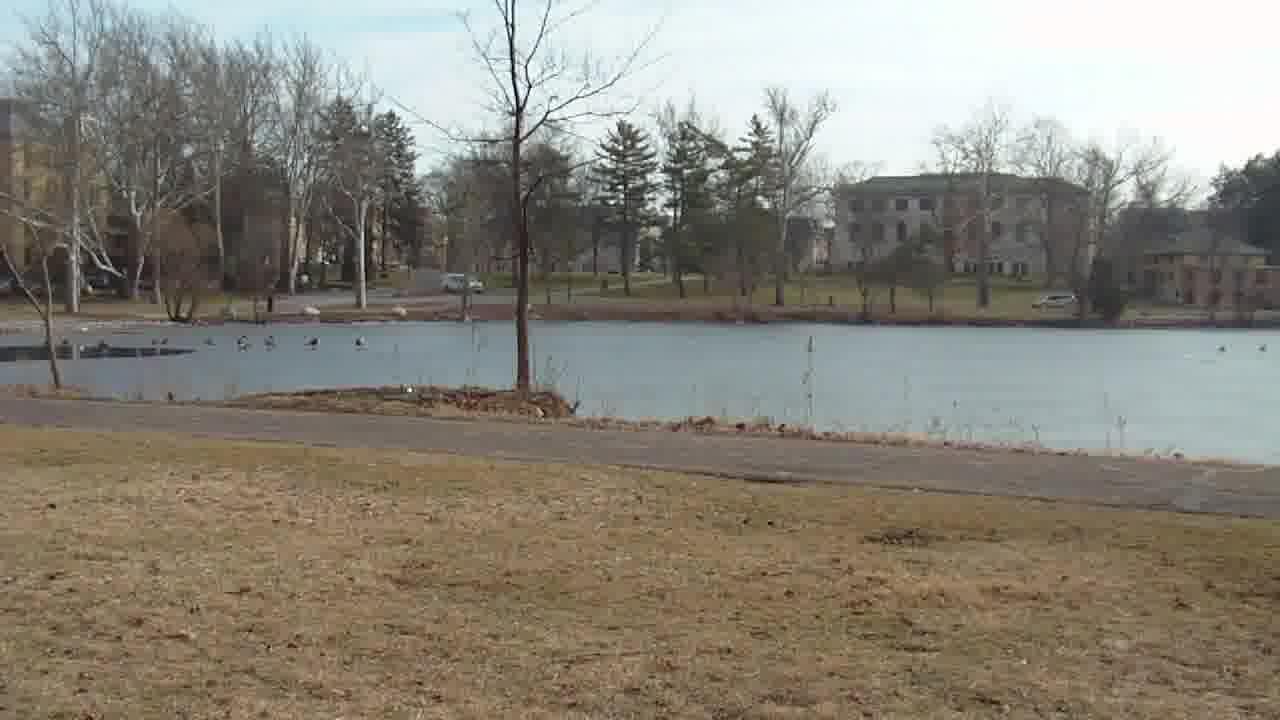}
        \caption{Controlled Ground Collection}
    \end{subfigure} \vspace{-2mm} 
    
    \caption{Examples of images in the three UG$^2$ collections.}
    \label{fig:datasets-samples}
    \vspace{-6mm}
\end{figure}

UG$^2$ is composed of $289$ videos with $1,217,496$ frames, representing $228$ ImageNet~\cite{ILSVRC15} classes extracted from annotated frames from three different video collections (see Supp. Table~2 for the complete list of classes). These classes are  further categorized into 37 super-classes encompassing visually similar ImageNet categories and two additional classes for pedestrian and resolution chart images (this distribution is explained in detail below). The three different video collections consist of: (1) 50 Creative Commons tagged videos taken by fixed-wing unmanned aerial vehicles (UAV) obtained from YouTube; (2) $61$ videos recorded by pilots of fixed wing gliders; and (3)  $178$ controlled videos captured on the ground specifically for this dataset.  Table~\ref{tab:dataset_summary} presents a summary of the dataset and Fig.~\ref{fig:datasets-samples} presents example frames from each of the collections. 

Furthermore, the dataset contains a subset of $159,464$ object-level annotated images.  Bounding boxes establishing object regions were manually determined using the Vatic tool for video annotation \cite{Vatic:2013}. Each annotation in the dataset indicates the position, scale, visibility and super-class for an object in a video. This is useful for running classification experiments. For example, for the baseline experiments described in Sec.~\ref{Results}, the objects were cropped out from the frames in a square region of at least $224\times224$ pixels (a common input size for many deep learning-based recognition models), using the annotations as a guide. Videos are also tagged to indicate problematic conditions.


\textbf{UAV Video Collection.} This collection found within UG$^2$ consists of video recorded from small UAVs in both rural and urban areas. The videos in this collection are open source content tagged with a Creative Commons license,  obtained from the YouTube video sharing site. Because of the source, they have different video resolutions (from $600\times400$ to $3840\times2026$) and frame rates (from $12$ FPS to $59$ FPS). This collection contains approximately $4$ hours of aerial video distributed across $50$ different videos. 

For this collection we observed 8 different video artifacts and other problems: glare/lens flare, poor image quality, occlusion, over/under exposure, camera shaking and noise (present in some videos that use autopilot telemetry), motion blur, and fish eye lens distortion. Additionally this collection contains videos with problematic weather/scene conditions such as night/low light video, fog, cloudy conditions and occlusion due to snowfall. Overall it contains $434,264$ frames. Across a subset of these frames we observed $31$ different super-classes (including the non-ImageNet pedestrians class), from which we extracted $32,608$ object images. The cropped object images have a diverse range of sizes, from $224\times224$ to $800\times800$.

\begin{figure}[t]
    \centering
    \includegraphics[width=0.45\textwidth]{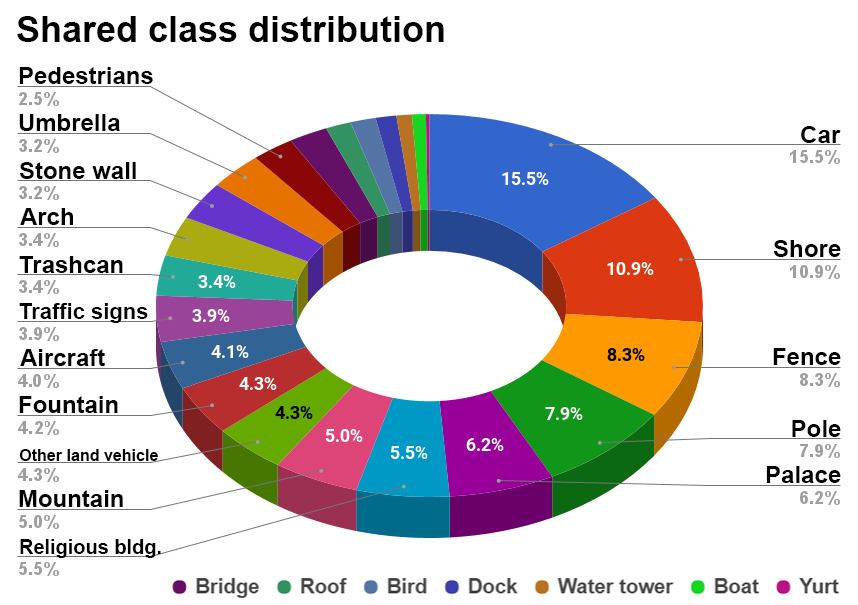}
    \caption{Distribution of annotated images belonging to classes shared by at least two different UG$^2$ collections.}
    \label{fig:sharedClassDistrib}
    \vspace{-6mm}
\end{figure}


\textbf{Glider Video Collection.} This collection found within UG$^2$ consists of video recorded by licensed pilots of fixed wing gliders in both rural and urban areas. It contains approximately $7$ hours of aerial video, distributed across $61$ different videos. The videos have frame rates ranging from $25$ FPS to $50$ FPS and different types of compression such as MTS, MP4 and MOV. 
Given the nature of this collection the videos mostly present imagery taken from thousands of feet above ground, further increasing the difficulty of object recognition tasks. Additionally, scenes of take off and landing contain artifacts such as motion blur, camera shaking, and occlusion (which in some cases is pervasive throughout the videos, showcasing parts of the glider that partially occlude the objects of interest). 

For the Glider Collection we observed $6$ different video artifacts and other problems: glare/lens flare, over/under exposure, camera shaking and noise, occlusion, motion blur, and fish eye lens distortion. Furthermore, this collection contains videos with problematic weather/scene conditions such as fog, clouds and occlusion due to rain.
Overall this collection contains $657,455$ frames. Across the annotated frames we observed $20$ different classes (including the non-ImageNet class of pedestrians), from which we extracted $31,760$ object images. The cropped object images have a diverse range of sizes, from $224\times224$ to $900\times900$.


\textbf{Ground Video Collection.} In order to provide some ground-truth with respect to problematic image conditions, we performed a video collection on the ground that intentionally  induced several common artifacts. One of the main challenges for object recognition within aerial images  is the difference in the scale of certain objects compared to those in the images used to train the recognition model. To address this, we recorded video of static objects  (\eg, flower pots, buildings) at a wide range of distances ($30$ft, $40$ft, $50$ft, $60$ft, $70$ft, $100$ft, $150$ft, and $200$ft). 

In conjunction with the differing recording distances, we induced motion blur in images using an orbital shaker to generate horizontal movement at different rotations per minute ($120$rpm, $140$rpm, $160$rpm, and $180$rpm). Parallel to this, we recorded video under different weather conditions (sun, clouds, rain, snow) that could affect object recognition, and employed a Sony Bloggie hand-held camera (with  $1280\times720$ resolution and a frame rate of 60 FPS) and a GoPro Hero 4 (with $1920\times1080$ resolution and a frame rate of 30 FPS), whose fisheye lens introduced further distortion. 

The Ground Collection contains approximately $40$ minutes of video, distributed across $178$ videos. Overall this collection represents $95,096$ annotated frames, and $28,009$ unannotated frames distributed across $42$ specific videos. The annotated frames contain $20$ different ImageNet classes. Furthermore, an additional class of videos showcasing a $9\times11$ inch $9\times9$ checkerboard grid exhibiting all aforementioned distances and all intervals of rotation. The motivation for including this artificial class is to provide a reference with well-defined  straight lines to assess the visual impact of image restoration and enhancement algorithms. The cropped object images have a diverse range of sizes: from $224\times224$ to $1000\times1000$.


\textbf{Object Categories and Distribution of Data.} A challenge presented by the objects annotated in the UAV and Glider collections is the high variability of both object scale and rotation. These two factors  make it difficult to differentiate some of the more fine-grained ImageNet categories. For example, while it may be easy to recognize a car from an aerial picture taken from hundreds (if not thousands) of feet above the ground, it might be impossible to determine whether that car is a taxi, a jeep or a sports car. An exception to this rule is the Ground Collection where we had more control over distances from the target which made possible the fine-grained class distinction. For example, chainlink-fence and bannisters are separate classes in the Ground Collection and are not combined to form a fence super-class. 


Thus UG$^2$ organizes the objects in high level classes that encompass multiple ImageNet synsets (ImageNet provides images for ``synsets" or ``synonym sets" of words or phrases that describe a concept in WordNet~\cite{fellbaum1998wordnet}; for more detail on the relationship between UG$^2$ and ImageNet classes see Supp. Table~1). Over 70\% of the UG$^2$ classes have more than 400 images and 58\% of the classes are present in the imagery of at least two collections (Fig.~\ref{fig:sharedClassDistrib}). Around 20\% of the classes are present in all three collections. \vspace{-3mm} 


\section{\vspace{-1.8mm}The UG$^2$ Classification Protocols} \label{ClassifMethod}

In order to establish good baselines for classification performance before and after the application of image enhancement and restoration algorithms, we used a selection of common deep learning approaches to recognize annotated objects and then considered the correct classification rate. Namely, we used the Keras~\cite{chollet2015keras} versions of the pre-trained networks VGG16 \& VGG19~\cite{VGG:2014}, Inception V3~\cite{Inception:2015}, and ResNet50~\cite{ResNet50:2015}. These experiments also serve as a demonstration of the UG$^2$ classification protocols. Each candidate restoration or enhancement algorithm should be treated as an image pre-processing step to prepare data to be submitted to all four networks, which serve as canonical classification references. The entirety of the annotated data for all three collections is used for evaluation, with the exceptions of the pedestrian and resolution chart classes, which do not belong to any synsets recognized by the networks. For our current experiments, we restricted our analysis on UG$^2$ dataset to pre-trained networks. Re-training the networks with our dataset would be considered in future. With respect to restoration and enhancement approaches that must be trained, we suggest a cross-dataset protocol [48] where some annotated training data should come from outside UG$^2$. However, un-annotated videos for additional validation purposes and parameter tuning are provided.

\textbf{Classification Metrics.} The networks used for the UG$^2$ classification task return a list of the ImageNet synsets along with the probability of the object belonging to each of the synsets classes. However, given what we discussed in Sec.~\ref{Dataset}, in some cases it is impossible to provide a fine-grained labeling for the annotated objects. Consequently, most of the super-classes we defined for UG$^2$ are composed of more than one ImageNet synset. That is, each annotated image \(i\) has a single super-class label \(L_i\) which in turn is defined by a set of ImageNet synsets \(L_i = \{s_1, ..., s_n\}\). 

To measure accuracy, we observe the number of correctly identified synsets in the top 5 predictions made by each pre-trained network. A prediction is considered to be correct if it's synset belongs to the set of synsets in the ground-truth super-class label. We use two metrics for this. The first measures the rate of detection of at least 1 correctly classified synset class. In other words, for a super-class label \(L_i = \{s_1, ..., s_n\}\), a network is able to detect 1 or more correctly classified synsets in the top 5 predictions. The second measures the rate of detecting all the possible correct synset classes in the super-class label synset set. For example, for a super-class label \(L_i = \{s_1, s_2, s_3\}\), a network is able to detect 3 correct synsets in the top 5 labels.\vspace{-4mm}

\section{\vspace{-1.8mm}Baseline Enhancement and Restoration} \label{Algorithms} 


To shed light on the effects image restoration and enhancement algorithms have on classification, we tested classic and state-of-the-art algorithms for image interpolation~\cite{Keys:1981:Interpolation}, super-resolution~\cite{Chao:2014:SRCNN, Kim:2016:VDSR}, and deblurring~\cite{Nah:2016:DSDD, Su:2016:DBN, Pan:2016:DCPD} (see Supp. Fig.~1 for examples). 


\textbf{Interpolation methods.} These classic methods attempt to obtain a high resolution image by up-sampling the source image (usually assuming the source image is a down-sampled version of the high resolution one) and by providing the best approximation of a pixel's color and intensity values depending on the nearby pixels. Since they do not need any prior training, they can be directly applied to any image. Nearest neighbor interpolation uses a weighted average of the nearby translated pixel values in order to calculate the output pixel value. Bilinear interpolation increases the number of translated pixel values to two and bicubic interpolation increases it to four.


\textbf{SRCNN.} The Super-Resolution Convolutional Neural Network (SRCNN)~\cite{Chao:2014:SRCNN} introduced deep learning techniques to super-resolution. The method employs a feedforward deep CNN to learn an end-to-end mapping between low resolution and high resolution images.
The network was trained on 5 million ``sub-images" generated from 395,909 images of the ILSVRC 2013 ImageNet detection training partition~\cite{ILSVRC15}. Typically, the results obtained from SRCNN can be distinguished from their low resolution counterparts by their sharper edges without visible artifacts. 

 
\textbf{VDSR.} The Very Deep Super Resolution (VDSR) algorithm~\cite{Kim:2016:VDSR} aims to outperform SRCNN  by employing a deeper CNN inspired by the VGG architecture~\cite{VGG:2014}. It also decreases training iterations and time by employing residual learning with a very high learning rate for faster convergence.
The VDSR network was trained on 291 images, collectively taken from Yang \etal \cite{yang2010image} and the Berkeley Segmentation Dataset \cite{martin2001database}. Unlike SRCNN, the network is capable of handling different scale factors. A good  image processed by VDSR is characterized by well-defined contours and a lack of edge effects at the borders. 



\textbf{Basic Blind Deconvolution.} The goal of any deblurring algorithm is to attempt to remove blur artifacts (\ie, the products of motion or depth variation, either from the object or the camera) that degrade image quality. 
This can be as simple as employing Matlab's blind deconvolution algorithm \cite{kundur1996blind}, which deconvolves the image using the maximum likelihood algorithm, with a $3\times3$ array of 1s as the initial point spread function. 

\textbf{Deep Video Deblurring.} The Deep Video Deblurring algorithm~\cite{Su:2016:DBN} was designed to address camera shake blur. However, in the results presented by Su \etal the algorithm also obtained good results for other types of blur, such as motion blur. This algorithm employs a CNN that was trained with video frames containing synthesized motion blur such that it receives a stack of neighboring frames and returns a deblurred frame. The algorithm allows for three types of frame-to-frame alignment: no alignment, optical flow alignment, and homography alignment. For our experiments we used optical flow alignment, which was reported to have the best performance with this algorithm.

\textbf{Deep Dynamic Scene Deblurring.} Similarly, the Deep Dynamic Scene Deblurring algorithm~\cite{Nah:2016:DSDD} utilizes deep learning in order to remove motion blur. Nah \etal implement a multi-scale CNN to restore blurred images in an end-to-end manner without assuming or estimating a blur kernel model. The network was trained using blurry images generated by averaging sequences (by considering gamma correction) of sharp frames in a dynamic scene with high speed cameras. Given that this algorithm was computationally expensive, we directly applied it to the cropped object regions, rather than to the full video frame.
\vspace{-3mm}
\section{\vspace{-1.8mm}UG$^2$ Baseline Results and Analysis} \label{Results}

\begin{figure}[t]
\centering
\scalebox{.75}{\pgfplotscreateplotcyclelist{my colors}{
        black!50\\
        red\\
        blue\\
        green\\
    }
    \pgfplotsset{
        compat=1.3,
        cycle list name=my colors,
        legend cell align=left,
    }

\begin{tikzpicture}
    \begin{axis}[
        title={Collection comparison: Rank 5 Classification},
        xlabel={At least 1 correct classification rate [\%]},
        ylabel={All possible classifications rate [\%]},
        ymin=0.0,
        minor y tick num=2,
        ymajorgrids=true,
        xmajorgrids=true,
        yminorgrids=true,
        minor grid style=loosely dotted,
        only marks,
        scatter,
        mark size=3.5pt,
        scatter src=explicit symbolic,
        table/meta=Method,
        scatter/classes={
            VGG16={mark=diamond*},
            VGG19={mark=halfdiamond*},
            Inception={mark=square*},
            ResNet={mark=triangle*}
        },
        legend entries={
            VGG16,
            VGG19,
            Inception,
            ResNet%
        },
        legend style={  at={(0.5,-0.2)},
                        anchor=north,legend columns=-1}
    ]

        \addplot table [
            x expr=-1,
            y expr=-1,
        ] {figures/YoutubeClassif.dat};
        
        \addplot table [
                x=Baseline-1C,
                y=Baseline-AC,
            ] {figures/YoutubeClassif.dat};
        \addplot table [
                x=Baseline-1C,
                y=Baseline-AC,
            ] {figures/KawaClassif.dat};
            \addplot table [
                x=Baseline-1C,
                y=Baseline-AC,
            ] {figures/GroundClassif.dat};

    \end{axis}

    \begin{axis}[
        xmin=1,
        xmax=2,
        ymin=1,
        ymax=2,
        hide axis,
        only marks,
        legend entries={
            ,       
            UAV Collection,
            Glider Collection,
            Ground Collection%
        },
        legend style={  at={(0.5,-0.2)},
                        anchor=north,
                        legend columns=3,
                        yshift=-20pt}
    ]
        \foreach \i in {0,...,3} {
            \addplot+ [mark=*] coordinates { (0,0) };
        }
    \end{axis}
\end{tikzpicture}}
\caption{Classification rates at rank 5 for the original, unprocessed, frames for each collection in the dataset.}
\label{fig:graph:Baseline_Comp}
\vspace{-4mm}
\end{figure}
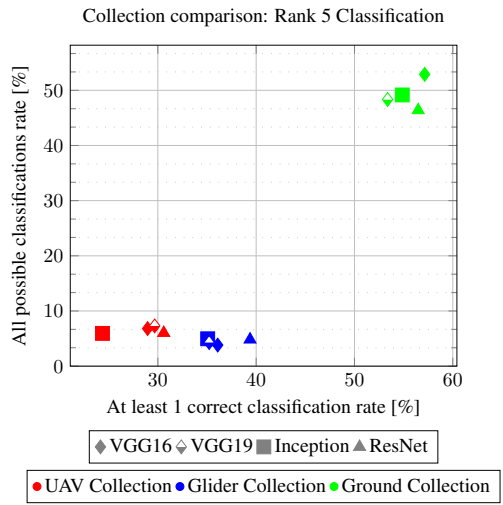

\textbf{Original Classification Results.} Fig.~\ref{fig:graph:Baseline_Comp} depicts the baseline classification results for the UG$^2$ collections, without any pre-processing, at rank 5 (results for top 1 predictions can be found in Supp. Figs.~2-4). Overall we observed distinct differences between the  results for all three collections, particularly between the airborne collections (UAV and Glider collections) and the Ground Collection.  These results establish that common deep learning networks alone cannot achieve good classification rates for this dataset.

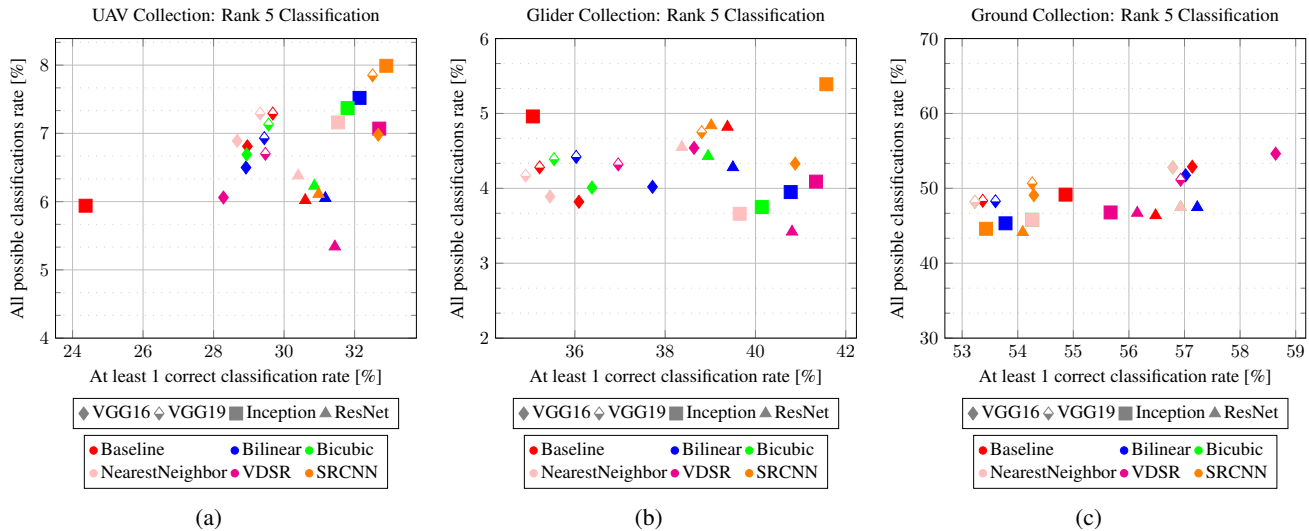
\begin{figure*}[!ht]
    \begin{subfigure}{0.33\textwidth}
        \scalebox{.7}{\pgfplotscreateplotcyclelist{my colors}{
        black!50\\
        red\\
        blue\\
        green\\
        pink\\
        magenta\\
        orange\\
    }
    \pgfplotsset{
        compat=1.3,
        cycle list name=my colors,
        legend cell align=left,
    }

\begin{tikzpicture}
    \begin{axis}[
        title={UAV Collection: Rank 5 Classification},
        xlabel={At least 1 correct classification rate [\%]},
        ylabel={All possible classifications rate [\%]},
        ymin=4.0,
        minor y tick num=2,
        ymajorgrids=true,
        xmajorgrids=true,
        yminorgrids=true,
        minor grid style=loosely dotted,
        only marks,
        scatter,
        mark size=3.5pt, 
        scatter src=explicit symbolic,
        table/meta=Method,
        scatter/classes={
            VGG16={mark=diamond*},
            VGG19={mark=halfdiamond*},
            Inception={mark=square*},
            ResNet={mark=triangle*}
        },
        legend entries={
            VGG16,
            VGG19,
            Inception,
            ResNet%
        },
        legend style={  at={(0.5,-0.2)},
                        anchor=north,legend columns=-1}
    ]

        \addplot table [
            x expr=0,
            y expr=0,
        ] {figures/YoutubeClassif.dat};

        \foreach \i in {
            Baseline,
            Bilinear,
            Bicubic,
            NearestNeighbor,
            VDSR,
            SRCNN%
        }{
            \addplot table [
                x=\i-1C,
                y=\i-AC,
            ] {figures/YoutubeClassif.dat};
        }

    \end{axis}

    \begin{axis}[
        xmin=1,
        xmax=2,
        ymin=1,
        ymax=2,
        hide axis,
        only marks,
        legend entries={
            ,       
            Baseline,
            Bilinear,
            Bicubic,
            NearestNeighbor,
            VDSR,
            SRCNN%
        },
        legend style={  at={(0.5,-0.2)},
                        anchor=north,
                        legend columns=3,
                        yshift=-20pt}
    ]
        \foreach \i in {0,...,6} {
            \addplot+ [mark=*] coordinates { (0,0) };
        }
    \end{axis}
\end{tikzpicture}}
        \caption{\vspace{-2.2mm}}
        \label{fig:classR5_interp_yt}
    \end{subfigure}
    \begin{subfigure}{0.33\textwidth}
        \scalebox{.7}{\pgfplotscreateplotcyclelist{my colors}{
        black!50\\
        red\\
        blue\\
        green\\
        pink\\
        magenta\\
        orange\\
    }
    \pgfplotsset{
        compat=1.3,
        cycle list name=my colors,
        legend cell align=left,
    }

\begin{tikzpicture}
    \begin{axis}[
        title={Glider Collection: Rank 5 Classification},
        xlabel={At least 1 correct classification rate [\%]},
        ylabel={All possible classifications rate [\%]},
        ymin=2.0, ymax=6.0,
        minor y tick num=2,
        ymajorgrids=true,
        xmajorgrids=true,
        yminorgrids=true,
        minor grid style=loosely dotted,
        only marks,
        scatter,
        mark size=3.5pt,
        scatter src=explicit symbolic,
        table/meta=Method,
        scatter/classes={
            VGG16={mark=diamond*},
            VGG19={mark=halfdiamond*},
            Inception={mark=square*},
            ResNet={mark=triangle*}
        },
        legend entries={
            VGG16,
            VGG19,
            Inception,
            ResNet%
        },
        legend style={  at={(0.5,-0.2)},
                        anchor=north,legend columns=-1}
    ]

        \addplot table [
            x expr=0,
            y expr=0,
        ] {figures/KawaClassif.dat};

        \foreach \i in {
            Baseline,
            Bilinear,
            Bicubic,
            NearestNeighbor,
            VDSR,
            SRCNN%
        }{
            \addplot table [
                x=\i-1C,
                y=\i-AC,
            ] {figures/KawaClassif.dat};
        }

    \end{axis}

    \begin{axis}[
        xmin=1,
        xmax=2,
        ymin=1,
        ymax=2,
        hide axis,
        only marks,
        legend entries={
            ,       
            Baseline,
            Bilinear,
            Bicubic,
            NearestNeighbor,
            VDSR,
            SRCNN%
        },
        legend style={  at={(0.5,-0.2)},
                        anchor=north,
                        legend columns=3,
                        yshift=-20pt}
    ]
        \foreach \i in {0,...,6} {
            \addplot+ [mark=*] coordinates { (0,0) };
        }
    \end{axis}
\end{tikzpicture}}
        \caption{\vspace{-2.2mm}}
        \label{fig:classR5_interp_glider}
    \end{subfigure}
    \begin{subfigure}{0.33\textwidth}
        \scalebox{.7}{\pgfplotscreateplotcyclelist{my colors}{
        black!50\\
        red\\
        blue\\
        green\\
        pink\\
        magenta\\
        orange\\
    }
    \pgfplotsset{
        compat=1.3,
        cycle list name=my colors,
        legend cell align=left,
    }

\begin{tikzpicture}
    \begin{axis}[
        title={Ground Collection: Rank 5 Classification},
        xlabel={At least 1 correct classification rate [\%]},
        ylabel={All possible classifications rate [\%]},
        ymin=30.0, ymax=70.0,
        minor y tick num=2,
        ymajorgrids=true,
        xmajorgrids=true,
        yminorgrids=true,
        minor grid style=loosely dotted,
        only marks,
        scatter,
        mark size=3.5pt,
        scatter src=explicit symbolic,
        table/meta=Method,
        scatter/classes={
            VGG16={mark=diamond*},
            VGG19={mark=halfdiamond*},
            Inception={mark=square*},
            ResNet={mark=triangle*}
        },
        legend entries={
            VGG16,
            VGG19,
            Inception,
            ResNet%
        },
        legend style={  at={(0.5,-0.2)},
                        anchor=north,legend columns=-1}
    ]

        \addplot table [
            x expr=0,
            y expr=0,
        ] {figures/GroundClassif.dat};

        \foreach \i in {
            Baseline,
            Bilinear,
            Bicubic,
            NearestNeighbor,
            VDSR,
            SRCNN%
        }{
            \addplot table [
                x=\i-1C,
                y=\i-AC,
            ] {figures/GroundClassif.dat};
        }

    \end{axis}

    \begin{axis}[
        xmin=1,
        xmax=2,
        ymin=1,
        ymax=2,
        hide axis,
        only marks,
        legend entries={
            ,       
            Baseline,
            Bilinear,
            Bicubic,
            NearestNeighbor,
            VDSR,
            SRCNN%
        },
        legend style={  at={(0.5,-0.2)},
                        anchor=north,
                        legend columns=3,
                        yshift=-20pt}
    ]
        \foreach \i in {0,...,6} {
            \addplot+ [mark=*] coordinates { (0,0) };
        }
    \end{axis}
\end{tikzpicture}}
        \caption{\vspace{-2.2mm}}
        \label{fig:classR5_interp_ground}
    \end{subfigure}
    \caption{\vspace{-2.2mm}Comparison of classification rates at rank 5 for each collection after applying resolution enhancement.}
    \label{fig:classR5_res}
        
\end{figure*}

Given the very poor quality of its videos, the UAV Collection turned out to be the most challenging in terms of object classification, obtaining the lowest classification performance out of the three collections. While the Glider Collection shared similar problematic conditions with the UAV Collection, we found that the images in this collection had a slightly higher classification rate than those in the UAV Collection in terms of identifying at least one correctly classified synset class. This improvement might be caused by the limited degree of movement of the gliders, since it ensured that the movement between frames was kept more stable over time, and by the camera's recording quality. The controlled Ground Collection yielded the highest classification rates, which, in an absolute sense, are still low.

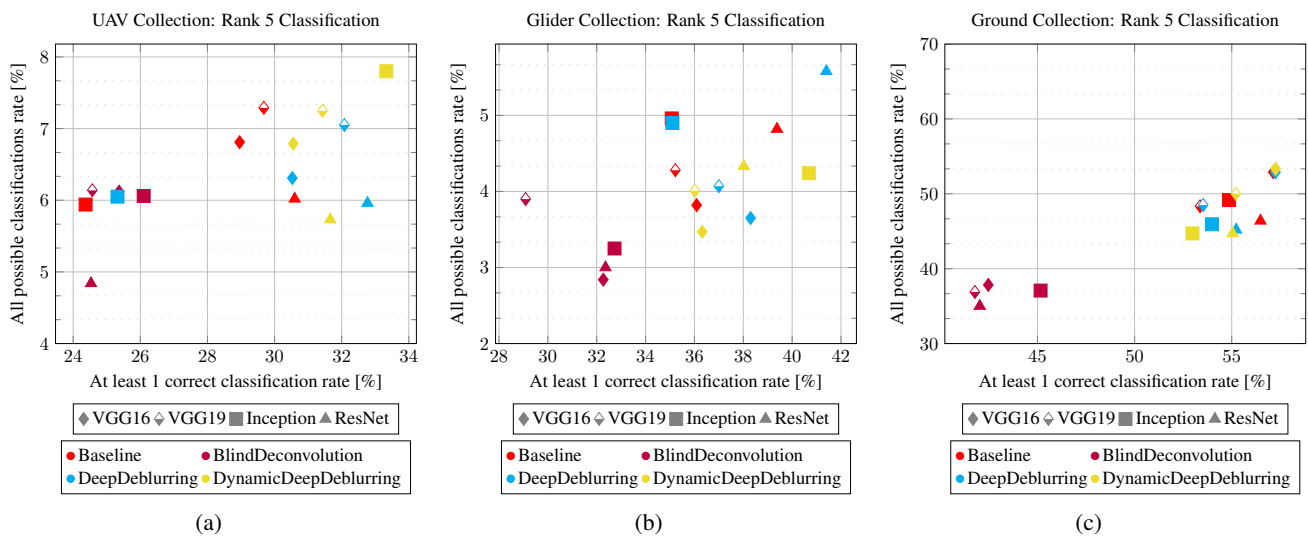
\begin{figure*}[ht]
    \begin{subfigure}{0.33\textwidth}
        \scalebox{.7}{\pgfplotscreateplotcyclelist{my colors}{
        black!50\\
        red\\
        purple\\
        cyan\\
        yellow!90!black\\
    }
    \pgfplotsset{
        compat=1.3,
        cycle list name=my colors,
        legend cell align=left,
    }

\begin{tikzpicture}
    \begin{axis}[
        title={UAV Collection: Rank 5 Classification},
        xlabel={At least 1 correct classification rate [\%]},
        ylabel={All possible classifications rate [\%]},
        ymin=4.0,
        minor y tick num=2,
        ymajorgrids=true,
        xmajorgrids=true,
        yminorgrids=true,
        minor grid style=loosely dotted,
        only marks,
        scatter,
        mark size=3.5pt,
        scatter src=explicit symbolic,
        table/meta=Method,
        scatter/classes={
            VGG16={mark=diamond*},
            VGG19={mark=halfdiamond*},
            Inception={mark=square*},
            ResNet={mark=triangle*}
        },
        legend entries={
            VGG16,
            VGG19,
            Inception,
            ResNet%
        },
        legend style={  at={(0.5,-0.2)},
                        anchor=north,legend columns=-1}
    ]

        \addplot table [
            x expr=0,
            y expr=0,
        ] {figures/YoutubeClassif.dat};

        \foreach \i in {
            Baseline,
            BlindDeconvolution,
            DeepDeblurring,
            DynamicDeepDeblurring%
        }{
            \addplot table [
                x=\i-1C,
                y=\i-AC,
            ] {figures/YoutubeClassif.dat};
        }

    \end{axis}

    \begin{axis}[
        xmin=1,
        xmax=2,
        ymin=1,
        ymax=2,
        hide axis,
        only marks,
        legend entries={
            ,       
            Baseline,
            BlindDeconvolution,
            DeepDeblurring,
            DynamicDeepDeblurring%
        },
        legend style={  at={(0.5,-0.2)},
                        anchor=north,
                        legend columns=2,
                        yshift=-20pt}
    ]
        \foreach \i in {0,...,6} {
            \addplot+ [mark=*] coordinates { (0,0) };
        }
    \end{axis}
\end{tikzpicture}}
        \caption{\vspace{-2.2mm}}
        \label{fig:classR5_debl_yt}
    \end{subfigure}
    \begin{subfigure}{0.33\textwidth}
        \scalebox{.7}{\pgfplotscreateplotcyclelist{my colors}{
        black!50\\
        red\\
        purple\\
        cyan\\
        yellow!90!black\\
    }
    \pgfplotsset{
        compat=1.3,
        cycle list name=my colors,
        legend cell align=left,
    }

\begin{tikzpicture}
    \begin{axis}[
        title={Glider Collection: Rank 5 Classification},
        xlabel={At least 1 correct classification rate [\%]},
        ylabel={All possible classifications rate [\%]},
        ymin=2.0,
        minor y tick num=2,
        ymajorgrids=true,
        xmajorgrids=true,
        yminorgrids=true,
        minor grid style=loosely dotted,
        only marks,
        scatter,
        mark size=3.5pt,
        scatter src=explicit symbolic,
        table/meta=Method,
        scatter/classes={
            VGG16={mark=diamond*},
            VGG19={mark=halfdiamond*},
            Inception={mark=square*},
            ResNet={mark=triangle*}
        },
        legend entries={
            VGG16,
            VGG19,
            Inception,
            ResNet%
        },
        legend style={  at={(0.5,-0.2)},
                        anchor=north,legend columns=-1}
    ]

        \addplot table [
            x expr=0,
            y expr=0,
        ] {figures/KawaClassif.dat};

        \foreach \i in {
            Baseline,
            BlindDeconvolution,
            DeepDeblurring,
            DynamicDeepDeblurring%
        }{
            \addplot table [
                x=\i-1C,
                y=\i-AC,
            ] {figures/KawaClassif.dat};
        }

    \end{axis}

    \begin{axis}[
        xmin=1,
        xmax=2,
        ymin=1,
        ymax=2,
        hide axis,
        only marks,
        legend entries={
            ,       
            Baseline,
            BlindDeconvolution,
            DeepDeblurring,
            DynamicDeepDeblurring%
        },
        legend style={  at={(0.5,-0.2)},
                        anchor=north,
                        legend columns=2,
                        yshift=-20pt}
    ]
        \foreach \i in {0,...,6} {
            \addplot+ [mark=*] coordinates { (0,0) };
        }
    \end{axis}
\end{tikzpicture}}
        \caption{\vspace{-2.2mm}}
        \label{fig:classR5_debl_glider}
    \end{subfigure}
    \begin{subfigure}{0.33\textwidth}
        \scalebox{.7}{\pgfplotscreateplotcyclelist{my colors}{
        black!50\\
        red\\
        purple\\
        cyan\\
        yellow!90!black\\
    }
    \pgfplotsset{
        compat=1.3,
        cycle list name=my colors,
        legend cell align=left,
    }

\begin{tikzpicture}
    \begin{axis}[
        title={Ground Collection: Rank 5 Classification},
        xlabel={At least 1 correct classification rate [\%]},
        ylabel={All possible classifications rate [\%]},
        ymin=30.0,ymax=70,
        minor y tick num=2,
        ymajorgrids=true,
        xmajorgrids=true,
        yminorgrids=true,
        minor grid style=loosely dotted,
        only marks,
        scatter,
        mark size=3.5pt,
        scatter src=explicit symbolic,
        table/meta=Method,
        scatter/classes={
            VGG16={mark=diamond*},
            VGG19={mark=halfdiamond*},
            Inception={mark=square*},
            ResNet={mark=triangle*}
        },
        legend entries={
            VGG16,
            VGG19,
            Inception,
            ResNet%
        },
        legend style={  at={(0.5,-0.2)},
                        anchor=north,legend columns=-1}
    ]

        \addplot table [
            x expr=0,
            y expr=0,
        ] {figures/GroundClassif.dat};

        \foreach \i in {
            Baseline,
            BlindDeconvolution,
            DeepDeblurring,
            DynamicDeepDeblurring%
        }{
            \addplot table [
                x=\i-1C,
                y=\i-AC,
            ] {figures/GroundClassif.dat};
        }

    \end{axis}

    \begin{axis}[
        xmin=1,
        xmax=2,
        ymin=1,
        ymax=2,
        hide axis,
        only marks,
        legend entries={
            ,       
            Baseline,
            BlindDeconvolution,
            DeepDeblurring,
            DynamicDeepDeblurring%
        },
        legend style={  at={(0.5,-0.2)},
                        anchor=north,
                        legend columns=2,
                        yshift=-20pt}
    ]
        \foreach \i in {0,...,4} {
            \addplot+ [mark=*] coordinates { (0,0) };
        }
    \end{axis}
\end{tikzpicture}}
        \caption{\vspace{-2.2mm}}
        \label{fig:classR5_debl_ground}
    \end{subfigure}
    \caption{Comparison of classification rates at rank 5 for each collection after applying deblurring.}\vspace{-3mm}
    \label{fig:classR5_debl}
\end{figure*}

\textbf{Effect of Restoration and Enhancement.} Ideally, image restoration and enhancement algorithms should help object recognition by improving poor quality images and should not impair it for good quality images. To test this assumption for the algorithms described in Sec.~\ref{Algorithms}, we used them to pre-process the annotated video frames of UG$^2$ and then proceeded to re-crop the objects of interest using the annotated bounding box information (as described in Sec.~\ref{Dataset}). 
Given that the scale of the images enhanced with the interpolation algorithms was doubled, the bounding boxes were scaled accordingly in those cases. Furthermore, the 
cropped object images were re-sized to $224\times224$ (input for VGG16, VGG19 and ResNet50) and $229\times229$ (input for Inception V3) during the classification experiments. See Supp. Tables~ 5-8, 9-13, and 13-16 for detailed breakdowns of the results for what follows.

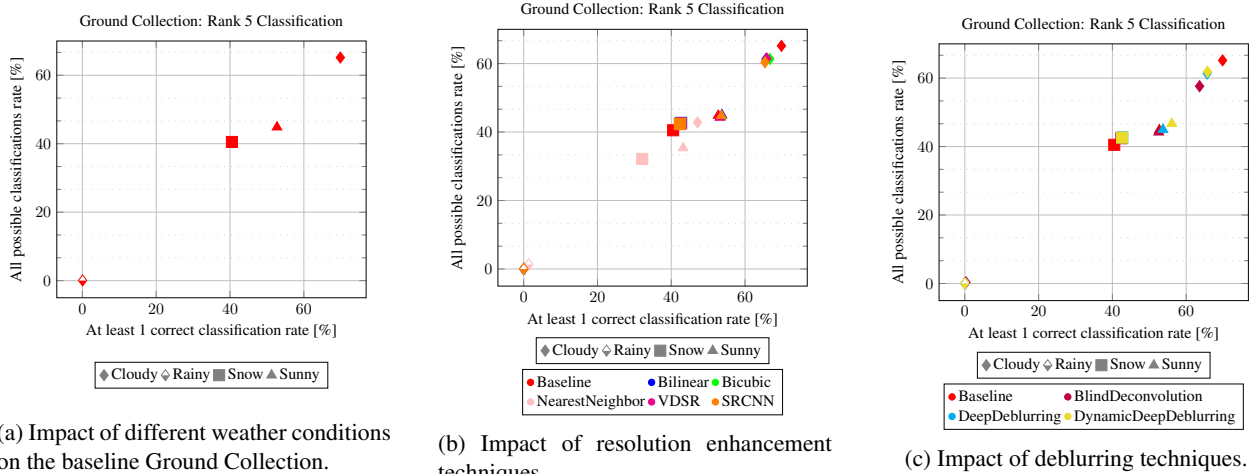
\begin{figure*}[t]
    \begin{subfigure}{0.3\textwidth}
        \scalebox{.6}{\pgfplotscreateplotcyclelist{my colors}{
        black!50\\
        red\\
        purple\\
        cyan\\
        yellow!90!black\\
    }
    \pgfplotsset{
        compat=1.3,
        cycle list name=my colors,
        legend cell align=left,
    }

\begin{tikzpicture}
    \begin{axis}[
        title={Ground Collection: Rank 5 Classification},
        xlabel={At least 1 correct classification rate [\%]},
        ylabel={All possible classifications rate [\%]},
        ymin=-5,ymax=70,
        minor y tick num=2,
        ymajorgrids=true,
        xmajorgrids=true,
        yminorgrids=true,
        minor grid style=loosely dotted,
        only marks,
        scatter,
        mark size=3.5pt,
        scatter src=explicit symbolic,
        table/meta=Method,
        scatter/classes={
            Cloudy={mark=diamond*},
            Rainy={mark=halfdiamond*},
            Snow={mark=square*},
            Sunny={mark=triangle*}
        },
        legend entries={
            Cloudy,
            Rainy,
            Snow,
            Sunny%
        },
        legend style={  at={(0.5,-0.25)},
                        anchor=north,legend columns=-1}
    ]

        \addplot table [
            x expr=-10,
            y expr=-10,
        ] {figures/GroundClassif_Weather.dat};

        \foreach \i in {
            Baseline%
        }{
            \addplot table [
                x=\i-1C,
                y=\i-AC,
            ] {figures/GroundClassif_Weather.dat};
        }

    \end{axis}

\end{tikzpicture}}
        \vspace{2mm}
        \caption{Impact of different weather conditions on the baseline Ground Collection.}
        \label{fig:classR5_weath_base}
    \end{subfigure} 
    \hspace{4.5mm}
    \begin{subfigure}{0.3\textwidth}
        \scalebox{.6}{\pgfplotscreateplotcyclelist{my colors}{
        black!50\\
        red\\
        blue\\
        green\\
        pink\\
        magenta\\
        orange\\
    }
    \pgfplotsset{
        compat=1.3,
        cycle list name=my colors,
        legend cell align=left,
    }

\begin{tikzpicture}
    \begin{axis}[
        title={Ground Collection: Rank 5 Classification},
        xlabel={At least 1 correct classification rate [\%]},
        ylabel={All possible classifications rate [\%]},
        ymin=-5, ymax=70.0,
        minor y tick num=2,
        ymajorgrids=true,
        xmajorgrids=true,
        yminorgrids=true,
        minor grid style=loosely dotted,
        only marks,
        scatter,
        mark size=3.5pt,
        scatter src=explicit symbolic,
        table/meta=Method,
        scatter/classes={
            Cloudy={mark=diamond*},
            Rainy={mark=halfdiamond*},
            Snow={mark=square*},
            Sunny={mark=triangle*}
        },
        legend entries={
            Cloudy,
            Rainy,
            Snow,
            Sunny%
        },
        legend style={  at={(0.5,-0.2)},
                        anchor=north,legend columns=-1}
    ]

        \addplot table [
            x expr=-10,
            y expr=-10,
        ] {figures/GroundClassif_Weather.dat};

        \foreach \i in {
            Baseline,
            Bilinear,
            Bicubic,
            NearestNeighbor,
            VDSR,
            SRCNN%
        }{
            \addplot table [
                x=\i-1C,
                y=\i-AC,
            ] {figures/GroundClassif_Weather.dat};
        }

    \end{axis}

    \begin{axis}[
        xmin=1,
        xmax=2,
        ymin=1,
        ymax=2,
        hide axis,
        only marks,
        legend entries={
            ,       
            Baseline,
            Bilinear,
            Bicubic,
            NearestNeighbor,
            VDSR,
            SRCNN%
        },
        legend style={  at={(0.5,-0.2)},
                        anchor=north,
                        legend columns=3,
                        yshift=-20pt}
    ]
        \foreach \i in {0,...,6} {
            \addplot+ [mark=*] coordinates { (0,0) };
        }
    \end{axis}
\end{tikzpicture}}
        \caption{Impact of resolution enhancement techniques.}
        \label{fig:classR5_weath_res}
    \end{subfigure} 
    \hspace{4.5mm}
    \begin{subfigure}{0.3\textwidth}
        \scalebox{.6}{\pgfplotscreateplotcyclelist{my colors}{
        black!50\\
        red\\
        purple\\
        cyan\\
        yellow!90!black\\
    }
    \pgfplotsset{
        compat=1.3,
        cycle list name=my colors,
        legend cell align=left,
    }

\begin{tikzpicture}
    \begin{axis}[
        title={Ground Collection: Rank 5 Classification},
        xlabel={At least 1 correct classification rate [\%]},
        ylabel={All possible classifications rate [\%]},
        ymin=-5,ymax=70,
        minor y tick num=2,
        ymajorgrids=true,
        xmajorgrids=true,
        yminorgrids=true,
        minor grid style=loosely dotted,
        only marks,
        scatter,
        mark size=3.5pt,
        scatter src=explicit symbolic,
        table/meta=Method,
        scatter/classes={
            Cloudy={mark=diamond*},
            Rainy={mark=halfdiamond*},
            Snow={mark=square*},
            Sunny={mark=triangle*}
        },
        legend entries={
            Cloudy,
            Rainy,
            Snow,
            Sunny%
        },
        legend style={  at={(0.5,-0.2)},
                        anchor=north,legend columns=-1}
    ]

        \addplot table [
            x expr=-10,
            y expr=-10,
        ] {figures/GroundClassif_Weather.dat};

        \foreach \i in {
            Baseline,
            BlindDeconvolution,
            DeepDeblurring,
            DynamicDeepDeblurring%
        }{
            \addplot table [
                x=\i-1C,
                y=\i-AC,
            ] {figures/GroundClassif_Weather.dat};
        }

    \end{axis}

    \begin{axis}[
        xmin=1,
        xmax=2,
        ymin=1,
        ymax=2,
        hide axis,
        only marks,
        legend entries={
            ,       
            Baseline,
            BlindDeconvolution,
            DeepDeblurring,
            DynamicDeepDeblurring%
        },
        legend style={  at={(0.5,-0.2)},
                        anchor=north,
                        legend columns=2,
                        yshift=-20pt}
    ]
        \foreach \i in {0,...,4} {
            \addplot+ [mark=*] coordinates { (0,0) };
        }
    \end{axis}
\end{tikzpicture}}
        \caption{Impact of deblurring techniques.}
        \label{fig:classR5_weath_deb}
    \end{subfigure}
    \caption{Comparison of classification rates for different weather conditions at rank 5 for the Ground Collection. To simplify the analysis, each point represents the performance when considering the output of all four networks simultaneously.}
    \label{fig:classR5_weath}
    \vspace{-6mm}
\end{figure*}

As can be observed in Figs.~\ref{fig:classR5_res} and~\ref{fig:classR5_debl}, the behaviour of the resolution enhancement and deblurring algorithms is different between the airborne and Ground collections. For the most part, both types of algorithms tended to improve the rate of identification for at least one correct class for all of the networks for the UAV and Glider collections (Figs.~\ref{fig:classR5_interp_yt},~\ref{fig:classR5_interp_glider},~\ref{fig:classR5_debl_yt},~and \ref{fig:classR5_debl_glider}). Over 60\% of the experiments reported an increase in the correct classification rate compared to that of the baseline. Conversely, for the Ground Collection, the restoration and enhancement algorithms seemed to impair the classification for all networks (Figs.~\ref{fig:classR5_interp_ground} and~\ref{fig:classR5_debl_ground}), going as far as reducing the at least one class identification performance by more than 16\% for some experiments. More than 60\% of the experiments reported a decrease in the classification rate for the Ground Collection. The property of hurting recognition performance on good quality imagery is certainly undesirable in these cases.

Further along these lines, while the classification rate for at least one correct class was increased for the airborne collections, after employing enhancement techniques the classification rate for finding all possible sub-classes in the super-class was negatively impacted for all three collections. Between 53\% and 68\% of the experiments reported a decrease in this metric. But this behaviour seemed to be more prevalent for the deblurring algorithms. For the UAV and Glider collections 75\% and 92\% of the deblurring experiments respectively had a negative impact in the classification rate for finding all possible classes, while only 40\% and 45\% of the resolution enhancement experiments reported a negative impact for the same metric.

We can also consider the performance with respect to individual networks. For the UAV  (Fig.~\ref{fig:classR5_interp_yt}) and Glider  (Fig.~\ref{fig:classR5_interp_glider}) Collections, SRCNN provided the best results for the VGG16 and VGG19 networks in both metrics, and was also the best in improving the rate of finding all the correct synsets of their respective super-class. Nevertheless, the best overall improvement for the rate of correctly classifying at least one class in both collections was achieved by employing the Dynamic Deep Deblurring algorithm, with an improvement of 8.96\% for the Inception network in the UAV case, and 6.5\% for the Glider case.
Resolution enhancement algorithms dominated the classification rate improvement for the Ground Collection, where VDSR obtained the highest improvement in both metrics for the VGG16, VGG19 (the best with 3.56\% improvement), and Inception networks, while Bilinear interpolation achieved the highest improvement for the ResNet50 network. 

In contrast, Blind Deconvolution drove down performance in all of the algorithms we tested for almost all networks. For the UAV Collection,  Blind Deconvolution  led to a decrease of at most 6.07\% in the rate of classifying at least 1 class correctly for the ResNet50 network. This behaviour was also observed for the Glider and Ground collections, where it led to the highest decreases in the classification rate of both metrics for all networks. These being 7\% for the ResNet50 network for the Glider Collection and 15.06\% for the VGG16 network for the Ground Collection. 



\textbf{Effect of Weather Conditions.} A significant contribution of our dataset is the availability of ground-truth for weather conditions in the Ground collection.
Without any pre-processing applied to that set, the classification performance under different weather conditions varies widely (Fig.~\ref{fig:classR5_weath_base}). In particular, there was a stark contrast between the classification rates of video taken during rainy weather and video taken under any other condition, with rain causing the classification rate for both metrics to drop. Likewise, snowy weather presented a lower classification rate than cloudy or sunny weather as it shares some of the problems of rainy video capture: adverse lighting conditions and partial object occlusion from the precipitation. Cloudy weather proved to be the most beneficial for image capture as those videos lacked most of the problems of the other conditions. Sunny conditions are not the best because of glare. This study confirms previously reported results for the impact of weather on recognition~\cite{Boult_2009_HRB}.

We also analyzed the interplay between the different restoration and enhancement algorithms and  different weather conditions (Figs.~\ref{fig:classR5_weath_res} and \ref{fig:classR5_weath_deb}; see Supp. Tables~17-20 for detailed results). For this analysis we observed that resolution enhancement algorithms provided small benefits for both metrics.  50\% of the experiments improved the correct classification rate of at least one class, and 40.63\%  improved the other metric. Again, resolution enhancement algorithms tended to provide the most improvement. The highest improvement (3.36\% for the correct classification rate of at least one class) was achieved for sunny weather by the VDSR algorithm.
Note that while classification for the more problematic weather conditions (rain, snow and sun) was improved, this was not the case for  cloudy weather, where the original images were already of high quality. \vspace{-2mm}



\section{\vspace{-1.8mm}Discussion} \label{Analysis}


The results of our experiments led to some surprises. While the restoration and enhancement algorithms tended to improve the classification results for the diverse imagery included in our dataset, no approach was able to improve the results by a significant margin. Moreover, in some cases, performance degraded after image pre-processing, particularly for higher quality frames, making these kind of pre-processing techniques unviable for heterogeneous datasets. We also noticed that different algorithms for the same type of image processing can have very different effects, as can different combinations of pre-processing and recognition algorithms. Depending on the metric considered, performance could be better or worse for various techniques. A possible reason for this can be that most of these networks were trained with images having a single type of image distortion and hence fail for images with multiple distortions from heterogenous sources. Significant improvement can be achieved if the networks are re-trained with UG$^2$. However, this needs further investigation and would be done in future. Thus, UG$^2$ dataset will prove to be useful for studying these phenomena for quite some time to come.

UG$^2$ forms the core of a large prize challenge that will be announced in Fall 2017 and run from  Spring to early Summer 2018. In this paper, we described one protocol that is part of that challenge. Several alternate protocols that are useful for research in this direction will also be included. For instance, we did not look at networks that combine feature learning, image enhancement and restoration, and classification. A protocol supporting this will be available. UG$^2$  can also be used for more traditional computational photography assessment (\ie, making the images look better), and this too will be supported. Stay tuned for more.

\textbf{Acknowledgement}
Funding was provided under IARPA contract \#2016-16070500002. This research is based upon work supported in part by the Office of the Director of National Intelligence (ODNI), Intelligence Advanced Research Projects Activity (IARPA). The views and conclusions contained herein are those of the authors and should not be interpreted as necessarily representing the official policies, either expressed or implied, of ODNI, IARPA, or the U.S. Government. The U.S. Government is authorized to reproduce and distribute reprints for governmental purposes notwithstanding any copyright annotation therein.
We thank Dr. Adam Czajka, visiting assistant professor at the University of Notre Dame and Mr. Sebastian Kawa for assistance with data collection.

{\small
\bibliographystyle{latex/ieee}
\bibliography{UAV_refs}

\begin{thebibliography}{10}\itemsep=-1pt

\bibitem{UCFAA}
{UCF Aerial Action} data set.
\newblock \url{http://crcv.ucf.edu/data/UCF_Aerial_Action.php}.

\bibitem{UCFARG}
{UCF-ARG} data set.
\newblock \url{http://crcv.ucf.edu/data/UCF-ARG.php}.

\bibitem{bevilacqua2012low}
M.~Bevilacqua, A.~Roumy, C.~Guillemot, and M.~L. Alberi-Morel.
\newblock Low-complexity single-image super-resolution based on nonnegative
  neighbor embedding.
\newblock In {\em British Machine Vision Conference}, 2012.

\bibitem{Boult_2009_HRB}
T.~E. Boult and W.~J. Scheirer.
\newblock Long range facial image acquisition and quality.
\newblock In M.~Tistarelli, S.~Li, and R.~Chellappa, editors, {\em In Handbook
  of Remote Biometrics: for Surveillance and Security (Springer-Verlag)}.
  Springer-Verlag, August 2009.

\bibitem{cho2009fast}
S.~Cho and S.~Lee.
\newblock Fast motion deblurring.
\newblock {\em ACM Transactions on Graphics (TOG)}, 28(5), 2009.

\bibitem{chollet2015keras}
F.~Chollet et~al.
\newblock Keras.
\newblock \url{https://github.com/fchollet/keras}, 2015.

\bibitem{Chao:2014:SRCNN}
C.~Dong, C.~C. Loy, K.~He, and X.~Tang.
\newblock Learning a deep convolutional network for image super-resolution.
\newblock In {\em European Conference on Computer Vision (ECCV)}, 2014.

\bibitem{efrat2013accurate}
N.~Efrat, D.~Glasner, A.~Apartsin, B.~Nadler, and A.~Levin.
\newblock Accurate blur models vs. image priors in single image
  super-resolution.
\newblock In {\em IEEE International Conference on Computer Vision (ICCV)},
  2013.

\bibitem{fellbaum1998wordnet}
C.~Fellbaum.
\newblock {\em WordNet: An Electronic Lexical Database}.
\newblock MIT Press, Cambridge, MA, 1998.

\bibitem{CAVIAR:Dataset:2004}
R.~B. Fisher.
\newblock The {PETS}04 surveillance ground-truth data sets.
\newblock In {\em Proc. 6th IEEE International Workshop on Performance
  Evaluation of Tracking and Surveillance}, 2004.

\bibitem{freedman2011image}
G.~Freedman and R.~Fattal.
\newblock Image and video upscaling from local self-examples.
\newblock {\em ACM Transactions on Graphics (TOG)}, 30(2), 2011.

\bibitem{freeman2002example}
W.~T. Freeman, T.~R. Jones, and E.~C. Pasztor.
\newblock Example-based super-resolution.
\newblock {\em IEEE Computer Graphics and Applications}, 22(2):56--65, 2002.

\bibitem{grgic2011scface}
M.~Grgic, K.~Delac, and S.~Grgic.
\newblock {SC}face--surveillance cameras face database.
\newblock {\em Multimedia Tools and Applications}, 51(3):863--879, 2011.

\bibitem{ResNet50:2015}
K.~He, X.~Zhang, S.~Ren, and J.~Sun.
\newblock Deep residual learning for image recognition.
\newblock {\em CoRR}, abs/1512.03385, 2015.

\bibitem{hennings2008simultaneous}
P.~H. Hennings-Yeomans, S.~Baker, and B.~V. Kumar.
\newblock Simultaneous super-resolution and feature extraction for recognition
  of low-resolution faces.
\newblock In {\em Computer Vision and Pattern Recognition (CVPR), 2008.} IEEE,
  2008.

\bibitem{huang2011super}
H.~Huang and H.~He.
\newblock Super-resolution method for face recognition using nonlinear mappings
  on coherent features.
\newblock {\em IEEE Transactions on Neural Networks}, 22(1):121--130, 2011.

\bibitem{huang2015single}
J.-B. Huang, A.~Singh, and N.~Ahuja.
\newblock Single image super-resolution from transformed self-exemplars.
\newblock In {\em IEEE Conference on Computer Vision and Pattern Recognition
  (CVPR), 2015.}, 2015.

\bibitem{jing2015super}
X.-Y. Jing, X.~Zhu, F.~Wu, X.~You, Q.~Liu, D.~Yue, R.~Hu, and B.~Xu.
\newblock Super-resolution person re-identification with semi-coupled low-rank
  discriminant dictionary learning.
\newblock In {\em IEEE Conference on Computer Vision and Pattern Recognition
  (CVPR), 2015.}, 2015.

\bibitem{joshi2009image}
N.~Joshi, C.~L. Zitnick, R.~Szeliski, and D.~J. Kriegman.
\newblock Image deblurring and denoising using color priors.
\newblock In {\em IEEE Conference on Computer Vision and Pattern Recognition
  (CVPR)}, 2009.

\bibitem{Keys:1981:Interpolation}
R.~Keys.
\newblock Cubic convolution interpolation for digital image processing.
\newblock {\em IEEE Transactions on Acoustics, Speech, and Signal Processing},
  29(6):1153--1160, Dec 1981.

\bibitem{Kim:2016:VDSR}
J.~Kim, J.~K. Lee, and K.~M. Lee.
\newblock Accurate image super-resolution using very deep convolutional
  networks.
\newblock In {\em IEEE Conference on Computer Vision and Pattern Recognition
  (CVPR)}, 2016.

\bibitem{kundur1996blind}
D.~Kundur and D.~Hatzinakos.
\newblock Blind image deconvolution.
\newblock {\em IEEE Signal Processing Magazine}, 13(3):43--64, 1996.

\bibitem{law2006lucky}
N.~M. Law, C.~D. Mackay, and J.~E. Baldwin.
\newblock Lucky imaging: high angular resolution imaging in the visible from
  the ground.
\newblock {\em Astronomy \& Astrophysics}, 446(2):739--745, 2006.

\bibitem{levin2007image}
A.~Levin, R.~Fergus, F.~Durand, and W.~T. Freeman.
\newblock Image and depth from a conventional camera with a coded aperture.
\newblock {\em ACM Transactions on Graphics (TOG)}, 26(3), 2007.

\bibitem{levin2011natural}
A.~Levin and B.~Nadler.
\newblock Natural image denoising: Optimality and inherent bounds.
\newblock In {\em IEEE Conference on Computer Vision and Pattern Recognition
  (CVPR)}, 2011.

\bibitem{levin2009understanding}
A.~Levin, Y.~Weiss, F.~Durand, and W.~T. Freeman.
\newblock Understanding and evaluating blind deconvolution algorithms.
\newblock In {\em IEEE Conference on Computer Vision and Pattern Recognition
  (CVPR)}, 2009.

\bibitem{levin2011efficient}
A.~Levin, Y.~Weiss, F.~Durand, and W.~T. Freeman.
\newblock Efficient marginal likelihood optimization in blind deconvolution.
\newblock In {\em IEEE Conference on Computer Vision and Pattern Recognition
  (CVPR)}, 2011.

\bibitem{lin2005face}
F.~Lin, J.~Cook, V.~Chandran, and S.~Sridharan.
\newblock Face recognition from super-resolved images.
\newblock In {\em Proceedings of the Eighth International Symposium on Signal
  Processing and Its Applications}, volume~2, 2005.

\bibitem{lin2007super}
F.~Lin, C.~Fookes, V.~Chandran, and S.~Sridharan.
\newblock Super-resolved faces for improved face recognition from surveillance
  video.
\newblock {\em Advances in Biometrics}, pages 1--10, 2007.

\bibitem{martin2001database}
D.~Martin, C.~Fowlkes, D.~Tal, and J.~Malik.
\newblock A database of human segmented natural images and its application to
  evaluating segmentation algorithms and measuring ecological statistics.
\newblock In {\em IEEE International Conference on Computer Vision (ICCV)},
  2001.

\bibitem{matsushita2006full}
Y.~Matsushita, E.~Ofek, W.~Ge, X.~Tang, and H.-Y. Shum.
\newblock Full-frame video stabilization with motion inpainting.
\newblock {\em IEEE Transactions on Pattern Analysis and Machine Intelligence
  (T-PAMI)}, 28(7):1150--1163, 2006.

\bibitem{mueller2016benchmark}
M.~Mueller, N.~Smith, and B.~Ghanem.
\newblock A benchmark and simulator for {UAV} tracking.
\newblock In {\em European Conference on Computer Vision (ECCV)}, 2016.

\bibitem{Nah:2016:DSDD}
S.~Nah, T.~H. Kim, and K.~M. Lee.
\newblock Deep multi-scale convolutional neural network for dynamic scene
  deblurring.
\newblock {\em CoRR}, abs/1612.02177, 2016.

\bibitem{nishiyama2009facial}
M.~Nishiyama, H.~Takeshima, J.~Shotton, T.~Kozakaya, and O.~Yamaguchi.
\newblock Facial deblur inference to improve recognition of blurred faces.
\newblock In {\em IEEE Conference on Computer Vision and Pattern Recognition
  (CVPR)}, 2009.

\bibitem{Virat:Dataset:2011}
S.~Oh, A.~Hoogs, A.~Perera, N.~Cuntoor, C.~C. Chen, J.~T. Lee, S.~Mukherjee,
  J.~K. Aggarwal, H.~Lee, L.~Davis, E.~Swears, X.~Wang, Q.~Ji, K.~Reddy,
  M.~Shah, C.~Vondrick, H.~Pirsiavash, D.~Ramanan, J.~Yuen, A.~Torralba,
  B.~Song, A.~Fong, A.~Roy-Chowdhury, and M.~Desai.
\newblock A large-scale benchmark dataset for event recognition in surveillance
  video.
\newblock In {\em IEEE Conference on Computer Vision and Pattern Recognition
  (CVPR)}, 2011.

\bibitem{Pan:2016:DCPD}
J.~Pan, D.~Sun, H.-P. Pfister, and M.-H. Yang.
\newblock Blind image deblurring using dark channel prior.
\newblock In {\em IEEE Conference on Computer Vision and Pattern Recognition
  (CVPR)}, 2016.

\bibitem{rasti2016convolutional}
P.~Rasti, T.~Uiboupin, S.~Escalera, and G.~Anbarjafari.
\newblock Convolutional neural network super resolution for face recognition in
  surveillance monitoring.
\newblock In {\em International Conference on Articulated Motion and Deformable
  Objects}, 2016.

\bibitem{reilly2013shadow}
V.~Reilly, B.~Solmaz, and M.~Shah.
\newblock Shadow casting out of plane (scoop) candidates for human and vehicle
  detection in aerial imagery.
\newblock {\em International Journal of Computer Vision (IJCV)},
  101(2):350--366, 2013.

\bibitem{ILSVRC15}
O.~Russakovsky, J.~Deng, H.~Su, J.~Krause, S.~Satheesh, S.~Ma, Z.~Huang,
  A.~Karpathy, A.~Khosla, M.~Bernstein, A.~C. Berg, and L.~Fei-Fei.
\newblock {ImageNet Large Scale Visual Recognition Challenge}.
\newblock {\em International Journal of Computer Vision (IJCV)},
  115(3):211--252, 2015.

\bibitem{CUHK:Dataset:2014}
J.~Shao, C.~C.~Loy, and X.~Wang.
\newblock Scene-independent group profiling in crowd.
\newblock In {\em IEEE Conference on Computer Vision and Pattern Recognition
  (CVPR)}, 2014.

\bibitem{sheikh2006statistical}
H.~R. Sheikh, M.~F. Sabir, and A.~C. Bovik.
\newblock A statistical evaluation of recent full reference image quality
  assessment algorithms.
\newblock {\em IEEE Transactions on Image Processing (T-IP)},
  15(11):3440--3451, 2006.

\bibitem{VGG:2014}
K.~Simonyan and A.~Zisserman.
\newblock Very deep convolutional networks for large-scale image recognition.
\newblock {\em CoRR}, abs/1409.1556, 2014.

\bibitem{Su:2016:DBN}
S.~Su, M.~Delbracio, J.~Wang, G.~Sapiro, W.~Heidrich, and O.~Wang.
\newblock Deep video deblurring.
\newblock {\em CoRR}, abs/1611.08387, 2016.

\bibitem{Inception:2015}
C.~Szegedy, V.~Vanhoucke, S.~Ioffe, J.~Shlens, and Z.~Wojna.
\newblock Rethinking the inception architecture for computer vision.
\newblock {\em CoRR}, abs/1512.00567, 2015.

\bibitem{timofte2013anchored}
R.~Timofte, V.~De~Smet, and L.~Van~Gool.
\newblock Anchored neighborhood regression for fast example-based
  super-resolution.
\newblock In {\em IEEE International Conference on Computer Vision (ICCV)},
  2013.

\bibitem{timofte2014}
R.~Timofte, V.~De~Smet, and L.~Van~Gool.
\newblock A+: Adjusted anchored neighborhood regression for fast
  super-resolution.
\newblock In {\em Asian Conference on Computer Vision (ACCV)}, 2014.

\bibitem{tomic2012toward}
T.~Tomic, K.~Schmid, P.~Lutz, A.~Domel, M.~Kassecker, E.~Mair, I.~L. Grixa,
  F.~Ruess, M.~Suppa, and D.~Burschka.
\newblock Toward a fully autonomous {UAV}: Research platform for indoor and
  outdoor urban search and rescue.
\newblock {\em IEEE Robotics \& Automation Magazine}, 19(3):46--56, 2012.

\bibitem{torralba2011unbiased}
A.~Torralba and A.~A. Efros.
\newblock Unbiased look at dataset bias.
\newblock In {\em IEEE Conference on Computer Vision and Pattern Recognition
  (CVPR)}, 2011.

\bibitem{uiboupin2016facial}
T.~Uiboupin, P.~Rasti, G.~Anbarjafari, and H.~Demirel.
\newblock Facial image super resolution using sparse representation for
  improving face recognition in surveillance monitoring.
\newblock In {\em 24th Signal Processing and Communication Application
  Conference (SIU)}, 2016.

\bibitem{Vatic:2013}
C.~Vondrick, D.~Patterson, and D.~Ramanan.
\newblock Efficiently scaling up crowdsourced video annotation.
\newblock {\em International Journal of Computer Vision (IJCV)},
  101(1):184--204, Jan. 2013.

\bibitem{wang2014recent}
R.~Wang and D.~Tao.
\newblock Recent progress in image deblurring.
\newblock {\em arXiv preprint arXiv:1409.6838}, 2014.

\bibitem{yang2014single}
C.-Y. Yang, C.~Ma, and M.-H. Yang.
\newblock Single-image super-resolution: A benchmark.
\newblock In {\em European Conference on Computer Vision (ECCV)}, 2014.

\bibitem{yang2008image}
J.~Yang, J.~Wright, T.~Huang, and Y.~Ma.
\newblock Image super-resolution as sparse representation of raw image patches.
\newblock In {\em IEEE Conference on Computer Vision and Pattern Recognition
  (CVPR)}, 2008.

\bibitem{yang2010image}
J.~Yang, J.~Wright, T.~S. Huang, and Y.~Ma.
\newblock Image super-resolution via sparse representation.
\newblock {\em IEEE Transactions on Image Processing (T-IP)},
  19(11):2861--2873, 2010.

\bibitem{LHI:Dataset:2007}
B.~Yao, X.~Yang, and S.-C. Zhu.
\newblock Introduction to a large-scale general purpose ground truth database:
  Methodology, annotation tool and benchmarks.
\newblock In {\em 6th International Conference on Energy Minimization Methods
  in Computer Vision and Pattern Recognition}, 2007.

\bibitem{yao2008improving}
Y.~Yao, B.~R. Abidi, N.~D. Kalka, N.~A. Schmid, and M.~A. Abidi.
\newblock Improving long range and high magnification face recognition:
  Database acquisition, evaluation, and enhancement.
\newblock {\em Computer Vision and Image Understanding (CVIU)},
  111(2):111--125, 2008.

\bibitem{yu2011face}
J.~Yu, B.~Bhanu, and N.~Thakoor.
\newblock Face recognition in video with closed-loop super-resolution.
\newblock In {\em IEEE Conference on Computer Vision and Pattern Recognition
  Workshops (CVPRW)}, 2011.

\bibitem{zeiler2010deconvolutional}
M.~D. Zeiler, D.~Krishnan, G.~W. Taylor, and R.~Fergus.
\newblock Deconvolutional networks.
\newblock In {\em IEEE Conference on Computer Vision and Pattern Recognition
  (CVPR)}, 2010.

\bibitem{zeiler2011adaptive}
M.~D. Zeiler, G.~W. Taylor, and R.~Fergus.
\newblock Adaptive deconvolutional networks for mid and high level feature
  learning.
\newblock In {\em IEEE International Conference on Computer Vision (ICCV)},
  2011.

\bibitem{zeyde2010single}
R.~Zeyde, M.~Elad, and M.~Protter.
\newblock On single image scale-up using sparse-representations.
\newblock In {\em International Conference on Curves and Surfaces}, 2010.

\bibitem{zhang2011close}
H.~Zhang, J.~Yang, Y.~Zhang, N.~M. Nasrabadi, and T.~S. Huang.
\newblock Close the loop: Joint blind image restoration and recognition with
  sparse representation prior.
\newblock In {\em IEEE International Conference on Computer Vision (ICCV)},
  2011.

\bibitem{TISI:Dataset:2013}
X.~Zhu, C.~C.~Loy, and S.~Gong.
\newblock Video synopsis by heterogeneous multi-source correlation.
\newblock In {\em IEEE International Conference on Computer Vision (ICCV)},
  2013.

\end{thebibliography}
}

\end{document}


\title{UG$^2$: a Video Benchmark for Assessing the Impact of Image Restoration and Enhancement on  Automatic Visual Recognition\\ Supplemental Material}

\maketitle
\ifwacvfinal\thispagestyle{empty}\fi

\section{Examples of enhancement algorithm results on UG$^2$}
\begin{figure}[!ht]
    \centering
    \begin{subfigure}{0.15\textwidth}
        \includegraphics[width=\textwidth]{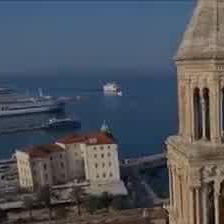}
        \caption{Baseline}
    \end{subfigure}
    \begin{subfigure}{0.15\textwidth}
        \includegraphics[width=\textwidth]{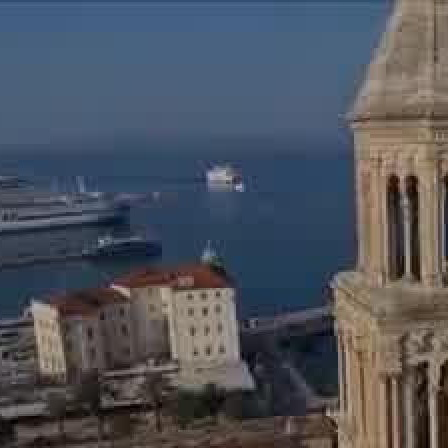}
        \caption{Nearest neighbor}
    \end{subfigure}
    \begin{subfigure}{0.15\textwidth}
        \includegraphics[width=\textwidth]{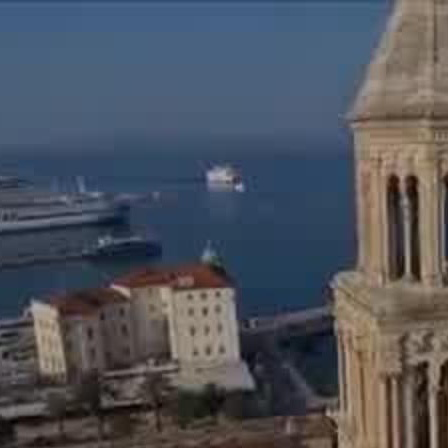}
        \caption{Bilinear}
    \end{subfigure}\hskip 5mm
    \begin{subfigure}{0.15\textwidth}
        \includegraphics[width=\textwidth]{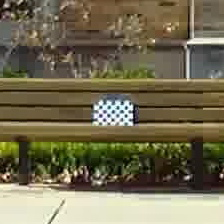}
        \caption{Baseline}
    \end{subfigure}
    \begin{subfigure}{0.15\textwidth}
        \includegraphics[width=\textwidth]{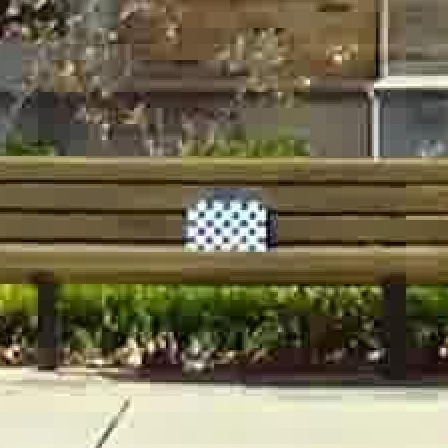}
        \caption{Nearest neighbor}
    \end{subfigure}
    \begin{subfigure}{0.15\textwidth}
        \includegraphics[width=\textwidth]{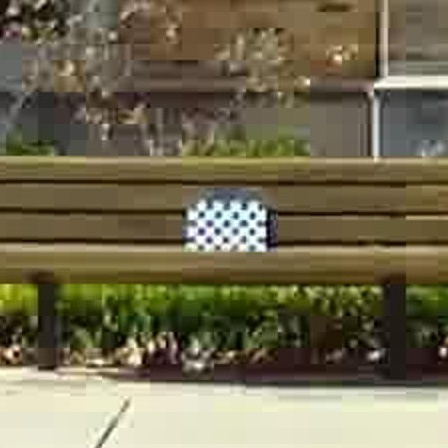}
        \caption{Bilinear}
    \end{subfigure}\vskip 2mm
    
    \begin{subfigure}{0.15\textwidth}
        \includegraphics[width=\textwidth]{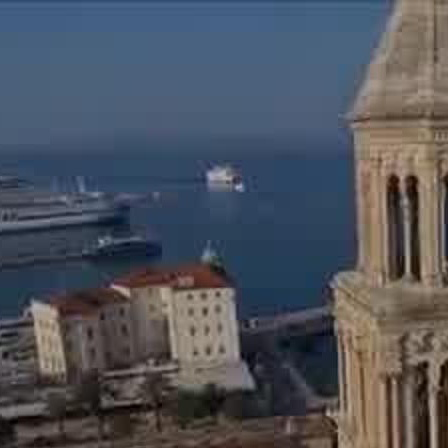}
        \caption{Bicubic}
    \end{subfigure}
    \begin{subfigure}{0.15\textwidth}
        \includegraphics[width=\textwidth]{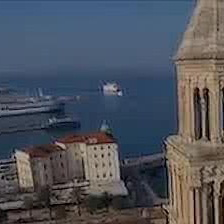}
        \caption{SRCNN}
    \end{subfigure}
    \begin{subfigure}{0.15\textwidth}
        \includegraphics[width=\textwidth]{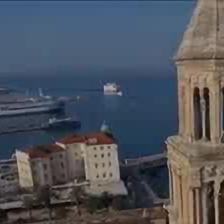}
        \caption{VDSR}
    \end{subfigure}\hskip 5mm
    \begin{subfigure}{0.15\textwidth}
        \includegraphics[width=\textwidth]{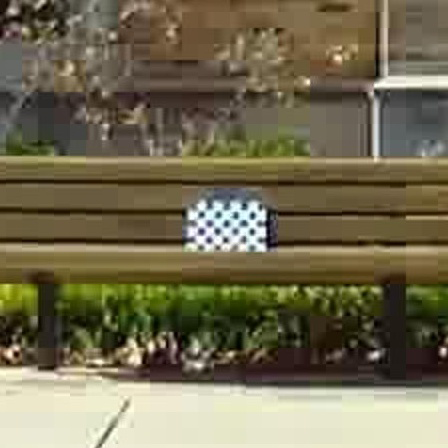}
        \caption{Bicubic}
    \end{subfigure}
    \begin{subfigure}{0.15\textwidth}
        \includegraphics[width=\textwidth]{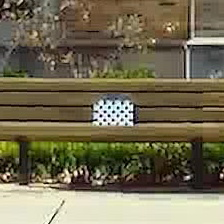}
        \caption{SRCNN}
    \end{subfigure}
    \begin{subfigure}{0.15\textwidth}
        \includegraphics[width=\textwidth]{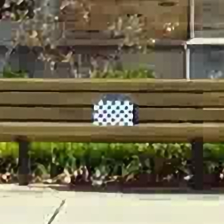}
        \caption{VDSR}
    \end{subfigure}\vskip 2mm

    \begin{subfigure}{0.15\textwidth}
        \includegraphics[width=\textwidth]{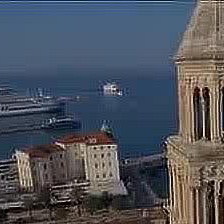}
        \caption{Blind deconv.}
    \end{subfigure}
    \begin{subfigure}{0.15\textwidth}
        \includegraphics[width=\textwidth]{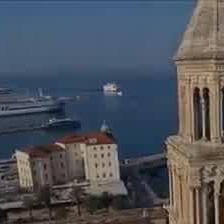}
        \caption{Deep deblurring}
    \end{subfigure}
    \begin{subfigure}{0.15\textwidth}
        \includegraphics[width=\textwidth]{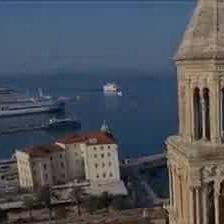}
        \caption{DDD}
    \end{subfigure} \hskip 5mm
    \begin{subfigure}{0.15\textwidth}
        \includegraphics[width=\textwidth]{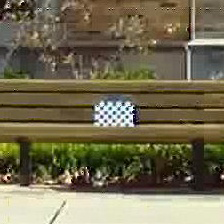}
        \caption{Blind deconv.}
    \end{subfigure}
    \begin{subfigure}{0.15\textwidth}
        \includegraphics[width=\textwidth]{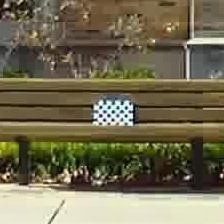}
        \caption{Deep deblurring}
    \end{subfigure}
    \begin{subfigure}{0.15\textwidth}
        \includegraphics[width=\textwidth]{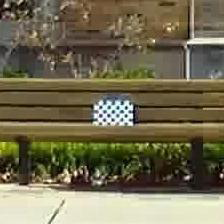}
        \caption{DDD}
    \end{subfigure}
    
    \caption{Examples of images processed with the methods analyzed. DDD stands for Dynamic Deep Deblurring. Best viewed in color and zoomed in.}
    \label{Supp-fig-retouched-samples}
\end{figure}

\pagebreak
\section{ImageNet and UG$^2$ equivalencies}

\begin{longtable}[c]{|p{.2\linewidth}|p{.2\linewidth}|p{.55\linewidth}|}
    \caption{Equivalencies between the UG$^2$ super-classes and ImageNet synsets. \textbf{(*)} These ImageNet classes are considered as super-classes for the Ground Collection classification, since we do have the fine-grained annotations for them. \textbf{(**)} These ImageNet classes are considered part of a super-class ``Bicycle" exclusive for the Ground Collection classification.}
    \label{Supp-tab:UG_INet_eqs}
    \hline
    \multicolumn{3}{| c |}{UG$^2$ - ImageNet equivalencies}\\
    \hline
    \textbf{UG2 class} & \textbf{Synset ID} & \textbf{ImageNet class} \\
    \hline
    \endfirsthead
    
    \hline
    \multicolumn{3}{|c|}{UG$^2$ - ImageNet equivalencies (Continuation)}\\
    \hline
    \textbf{UG2 class} & \textbf{Synset ID} & \textbf{ImageNet class} \\
    \hline
    \endhead
    
    \endfoot
    
    \hline
    
    \endlastfoot

    Aircraft  & n02690373 & airliner \\
 & n04266014 & space shuttle \\
 & n04552348 & warplane, military plane \\
 & n04592741 & wing \\
\hline
AnalogClock & n02708093 & analog clock \\
\hline
Arch & n04486054 & triumphal arch \\
\hline
Bird & n01817953 & African grey, African gray, Psittacus erithacus \\
 & n02018207 & American coot, marsh hen, mud hen, water hen, Fulica americana \\
 & n02009912 & American egret, great white heron, Egretta albus \\
 & n01828970 & bee eater \\
 & n02011460 & bittern \\
 & n01795545 & black grouse \\
 & n01530575 & brambling, Fringilla montifringilla \\
 & n01560419 & bulbul \\
 & n02018795 & bustard \\
 & n01592084 & chickadee \\
 & n01824575 & coucal \\
 & n02033041 & dowitcher \\
 & n01847000 & drake \\
 & n02017213 & European gallinule, Porphyrio porphyrio \\
 & n01531178 & goldfinch, Carduelis carduelis \\
 & n01855672 & goose \\
 & n01829413 & hornbill \\
 & n01537544 & indigo bunting, indigo finch, indigo , Passerina cyanea \\
 & n01843065 & jacamar \\
 & n01580077 & jay \\
 & n01608432 & kite \\
 & n02013706 & limpkin, Aramus pictus \\
 & n02009229 & little blue heron, Egretta caerulea \\
 & n01582220 & magpie \\
 & n02037110 & oystercatcher, oyster catcher \\
 & n01807496 & partridge \\
 & n01796340 & ptarmigan \\
 & n02027492 & red-backed sandpiper, dunlin, Erolia alpina \\
 & n01855032 & red-breasted merganser, Mergus serrator \\
 & n02028035 & redshank, Tringa totanus \\
 & n01558993 & robin, American robin, Turdus migratorius \\
 & n02025239 & ruddy turnstone, Arenaria interpres \\
 & n01797886 & ruffed grouse, partridge, Bonasa umbellus \\
 & n02006656 & spoonbill \\
 & n01819313 & sulphur-crested cockatoo, Kakatoe galerita, Cacatua galerita \\
 & n01601694 & water ouzel, dipper \\
\hline
Boat & n02687172 & aircraft carrier, carrier, flattop, attack aircraft carrier \\
 & n02951358 & canoe \\
 & n02981792 & catamaran \\
 & n03095699 & container ship, containership, container vessel \\
 & n03344393 & fireboat \\
 & n03447447 & gondola \\
 & n03662601 & lifeboat \\
 & n03673027 & liner, ocean liner \\
 & n03947888 & pirate, pirate ship \\
 & n04147183 & schooner \\
 & n04273569 & speedboat \\
 & n04483307 & trimaran \\
 & n04606251 & wreck \\
 & n04612504 & yawl \\
\hline
Bridge & n03933933 & pier \\
 & n04311004 & steel arch bridge \\
 & n04366367 & suspension bridge \\
\hline
Car & n02701002 & ambulance \\
& n02704792 & amphibian, amphibious vehicle \\
& n02814533 & beach wagon, station wagon, wagon, estate car, beach waggon, station waggon, waggon \\
& n02930766 & cab, hack, taxi, taxicab \\
& n03100240 & convertible \\
& n03345487 & fire engine, fire truck \\
& n03417042 & garbage truck, dustcart \\
& n03444034 & go-kart \\
& \textbf{n03445924*} & \textbf{golfcart, golf cart} \\
& n03594945 & jeep, landrover \\
& n03670208 & limousine, limo \\
& n03769881 & minibus \\
& n03770679 & minivan \\
& n03776460 & mobile home, manufactured home \\
& n03777568 & Model T \\
& n03796401 & moving van \\
& n03895866 & passenger car, coach, carriage \\
& n03930630 & pickup, pickup truck \\
& n03977966 & police van, police wagon, paddy wagon, patrol wagon, wagon, black Maria \\
& n04037443 & racer, race car, racing car \\
& n04065272 & recreational vehicle, RV, R.V. \\
& n04146614 & school bus \\
& n04285008 & sports car, sport car \\
& n04335435 & streetcar, tram, tramcar, trolley, trolley car \\
& n04461696 & tow truck, tow car, wrecker \\
& n04467665 & trailer truck, tractor trailer, trucking rig, rig, articulated lorry, semi \\
& n04487081 & trolleybus, trolley coach, trackless trolley \\
\hline
Mountain & n09193705 & alp \\
 & n09246464 & cliff, drop, drop-off \\
 & n09399592 & promontory, headland, head, foreland \\
 & n09468604 & valley, vale \\
 & n09472597 & volcano \\
\hline
ConstructionMachinery & n02012849 & crane \\
 & n03384352 & forklift \\
 & n03532672 & hook, claw \\
 & n04252225 & snowplow, snowplough \\
\hline
Dock & n02859443 & boathouse \\
 & n03216828 & dock, dockage, docking facility \\
\hline
Dog & n02116738 & African hunting dog, hyena dog, Cape hunting dog, Lycaon pictus \\
& n02096294 & Australian terrier \\
& n02110806 & basenji \\
& n02106166 & Border collie \\
& n02106382 & Bouvier des Flandres, Bouviers des Flandres \\
& n02108089 & boxer \\
& n02101388 & Brittany spaniel \\
& n02114855 & coyote, prairie wolf, brush wolf, Canis latrans \\
& n02110341 & dalmatian, coach dog, carriage dog \\
& n02107142 & Doberman, Doberman pinscher \\
& n02100735 & English setter \\
& n02109961 & Eskimo dog, husky \\
& n02099267 & flat-coated retriever \\
& n02106662 & German shepherd, German shepherd dog, German police dog, alsatian \\
& n02099601 & golden retriever \\
& n02109047 & Great Dane \\
& n02113712 & miniature poodle \\
& n02111277 & Newfoundland, Newfoundland dog \\
& n02086079 & Pekinese, Pekingese, Peke \\
& n02114712 & red wolf, maned wolf, Canis rufus, Canis niger \\
& n02087394 & Rhodesian ridgeback \\
& n02092002 & Scottish deerhound, deerhound \\
& n02093256 & Staffordshire bullterrier, Staffordshire bull terrier \\
& n02113799 & standard poodle \\
& n02108551 & Tibetan mastiff \\
& n02114367 & timber wolf, grey wolf, gray wolf, Canis lupus \\
& n02100583 & vizsla, Hungarian pointer \\
& n02114548 & white wolf, Arctic wolf, Canis lupus tundrarum \\
& n02094433 & Yorkshire terrier \\
\hline
FarmMachines & n03496892 & harvester, reaper \\
 & n03967562 & plow, plough \\
 & n04428191 & thresher, thrasher, threshing machine \\
 & n04465501 & tractor \\
\hline
Fence & \textbf{n02788148*} & \textbf{bannister, banister, balustrade, balusters, handrail} \\
 & \textbf{n03000134*} & \textbf{chainlink fence} \\
 & n03930313 & picket fence, paling \\
 & n04604644 & worm fence, snake fence, snake-rail fence, Virginia fence \\
\hline
FlowerPot & n03991062 & pot, flowerpot \\
\hline
Fountain & \textbf{n03388043*} & \textbf{fountain} \\
 & n03903868 & pedestal, plinth, footstall \\
\hline
Greenhouse & n03457902 & greenhouse, nursery, glasshouse \\
\hline
Lighthouse & n02814860 & beacon, lighthouse, beacon light, pharos \\
\hline
Livestock & n02415577 & bighorn, bighorn sheep, cimarron, Rocky Mountain bighorn, Rocky Mountain sheep, Ovis canadensis \\
 & n02410509 & bison \\
 & n02395406 & hog, pig, grunter, squealer, Sus scrofa \\
 & n02437616 & llama \\
 & n02403003 & ox \\
 & n02412080 & ram, tup \\
 & n02389026 & sorrel \\
 & n02397096 & warthog \\
\hline
Obelisk & n03837869 & obelisk \\
 & n04458633 & totem pole \\
\hline
OtherLandVehicle & \textbf{n02835271**} & \textbf{bicycle-built-for-two, tandem bicycle, tandem} \\
 & n02860847 & bobsled, bobsleigh, bob \\
 & n03218198 & dogsled, dog sled, dog sleigh \\
 & n03538406 & horse cart, horse-cart \\
 & n03785016 & moped \\
 & n03791053 & motor scooter, scooter \\
 & \textbf{n03792782**} & \textbf{mountain bike, all-terrain bike, off-roader} \\
 & n03868242 & oxcart \\
 & n04252077 & snowmobile \\
 & n04482393 & tricycle, trike, velocipede \\
 & n02917067 & bullet train, bullet \\
 & n03272562 & electric locomotive \\
 & n03393912 & freight car \\
 & n04310018 & steam locomotive \\
\hline
Palace & n02980441 & castle \\
 & \textbf{n03220513*} & \textbf{dome} \\
 & n03788195 & mosque \\
 & n03877845 & palace \\
 & n03956157 & planetarium \\
 & n04346328 & stupa, tope \\
\hline
Parachute & n03888257 & parachute, chute \\
\hline
ParkBench & n03891251 & park bench \\
\hline
Patio & n03899768 & patio, terrace \\
\hline
Pole & \textbf{n03355925*} & \textbf{flagpole, flagstaff} \\
 & \textbf{n03976657*} & \textbf{pole} \\
\hline
ReligiousBuilding & n03028079 & church, church building \\
 & n03781244 & monastery \\
\hline
Roof & n04417672 & thatch, thatched roof \\
 & n04435653 & tile roof \\
\hline
Ruins & n03042490 & cliff dwelling \\
\hline
Shore & n09332890 & lakeside, lakeshore \\
 & n09421951 & sandbar, sand bar \\
 & n09428293 & seashore, coast, seacoast, sea-coast \\
\hline
StoneWall & n02894605 & breakwater, groin, groyne, mole, bulwark, seawall, jetty \\
 & n04326547 & stone wall \\
\hline
Swing & n04371774 & swing \\
\hline
Tent & n03792972 & mountain tent \\
\hline
Trashcan & n02747177 & ashcan, trash can, garbage can, wastebin, ash bin, ash-bin, ashbin, dustbin, trash barrel, trash bin \\
\hline
StreetSigns  & n06794110 & street sign \\
 & n06874185 & traffic light, traffic signal, stoplight \\
\hline
Umbrella & n04507155 & umbrella \\
\hline
WaterTower & n04562935 & water tower \\
\hline
Yurt & n04613696 & yurt
\end{longtable}

\pagebreak
\section{UG$^2$ Collection's details}

~
\begin{table}[H]
\caption{Number of images per class in each collection}
\label{Supp-tab:Imgs_per_class}
\begin{tabular}{|c|c|c|}
\hline
\textbf{UG$^2$ super-class} & \textbf{UAV Collection}  & \textbf{Glider Collection} \\ 
\hline\hline
Aircraft & 1240 & 4099 \\
AnalogClock & 0 & 0 \\
Arch & 84 & 0 \\
Bird & 1105 & 877 \\
Boat & 161 & 921 \\
Bridge & 401 & 2630 \\
Car & 5790 & 4811 \\
Mountain & 2210 & 4436 \\
ConstructionMachinery & 401 & 0 \\
Dock & 755 & 857 \\
Dog & 299 & 0 \\
FarmMachines & 185 & 0 \\
Fence & 462 & 790 \\
FlowerPot & 0 & 0 \\
Fountain & 0 & 414 \\
Greenhouse & 0 & 292 \\
Lighthouse & 74 & 0 \\
Livestock & 848 & 0 \\
Obelisk & 306 & 0 \\
OtherLandVehicle & 369 & 480 \\
Palace & 608 & 2035 \\
Parachute & 134 & 0 \\
ParkBench & 0 & 0 \\
Patio & 0 & 0 \\
Pedestrians & 2782 & 515 \\
Pole & 2661 & 871 \\
ReligiousBuilding & 2265 & 327 \\
ResolutionChart & 0 & 0 \\
Roof & 743 & 1240 \\
Ruins & 1654 & 0 \\
Seashore & 4250 & 5732 \\
StoneWall & 60 & 0 \\
Swing & 52 & 0 \\
Tent & 85 & 0 \\
StreetSigns & 1006 & 0 \\
Trashcan & 355 & 77 \\
Umbrella & 154 & 0 \\
WaterTower & 883 & 314 \\
Yurt & 226 & 42 \\
\hline
\end{tabular}
\quad
\begin{tabular}{|c|c|}
\hline
\textbf{UG$^2$ super-class} & \textbf{Ground Collection} \\ 
\hline\hline
AnalogClock & 6347 \\
Bannister & 3958 \\
BasilicaChurch & 4647 \\
Bicycle & 4793 \\
Car & 6654 \\
ChainlinkFence & 5697 \\
Dome & 5567 \\
Flagpost & 5273 \\
FlowerPot & 5530 \\
Fountain & 5187 \\
Golfcart & 3148 \\
Lakeshore & 4371 \\
Lamppost & 1642 \\
ParkBench & 4946 \\
Patio & 5757 \\
ResolutionChart & 3478 \\
StoneWall & 4220 \\
StreetSign & 4108 \\
Trashcan & 4047 \\
TriumphalArch & 4350 \\
Umbrella & 4048 \\
\hline
\end{tabular}
\quad
\end{table}

~
\begin{table}[H]
\caption{Number of videos with problematic conditions in the airborne collections} 
\label{Supp-tab:Airborne_artifacts}
\begin{center}
\begin{tabular}{|c|c|c|}
\hline
\textbf{Problematic condition} & \textbf{UAV Collection}  & \textbf{Glider Collection} \\
\hline\hline

Camera blur (dirty) & 4 & 16 \\
Fish eye & 1 & 0 \\
Fog/Clouds & 1 & 13 \\
Mild annotations & 1 & 0 \\
Mild glare/flare & 3 & 1 \\
Mild noise & 3 & 0 \\
Mild shaking & 2 & 14 \\
Minor annotations & 3 & 0 \\
Minor glare/flare & 10 & 11 \\
Minor noise & 2 & 4 \\
Minor shaking & 5 & 10 \\
Motion blur & 2 & 2 \\
Night video/B\&W & 3 & 0 \\
Averse light conditions & 12 & 8 \\
Poor image quality & 10 & 1 \\
Prevalent annotations & 8 & 0 \\
Prevalent noise & 5 & 0 \\
Prevalent shaking & 1 & 0 \\
Snow & 2 & 0 \\
Rain & 0 & 2 \\

\hline
\end{tabular}
\end{center}
\end{table}

~
\begin{table}[H]
\caption{Number of images with controlled and uncontrolled problematic conditions in the Ground collection} \label{Supp-tab:Ground_artifacts}
\begin{tabular}{|c|c|}
\hline
\textbf{Distance (ft)} & \textbf{Total images}\\
\hline\hline
100 & 8,905\\
150 & 10,766\\
200 & 6,535\\
50 & 10,914\\
60 & 7,942\\
70 & 10,673\\
30 & 34,233\\
40 & 8,606\\
\hline
\end{tabular}
\quad
\begin{tabular}{|c|c|}
\hline
\textbf{Movement (RPM)} & \textbf{Total images}\\
\hline\hline
NA & 63,503\\
120 & 8,513\\
140 & 9,170\\
160 & 9,235\\
180 & 8,153\\
100 & 98,574\\
\hline
\end{tabular}
\quad
\begin{tabular}{|c|c|}
\hline
\textbf{Weather} &  \textbf{Total images}\\
\hline\hline
Sunny & 64,942\\
Snow & 9,866\\
Cloudy & 23,376\\
Rainy & 390\\
\hline
\end{tabular}
\quad
\begin{tabular}{|c|c|}
\hline
\textbf{Camera} &   \textbf{Total Images}\\
\hline\hline
Go pro & 20,808\\
Sony & 77,766\\
\hline
\end{tabular}
\quad
\end{table}

\pagebreak
\section{Rank 5 classification details}

\textbf{1C} stands for the rate of correct classification of at least one class

\textbf{AC} stands for the rate of correct classification of all possible classes

\subsection{UAV Collection}

\begin{table}[H]
\vspace{-3mm}
 \caption{Details for the UAV Collection's classification results at rank 5} \label{Supp-tab:YT_t5_overall_dets}
\vspace{-3mm}
\begin{center}
\begin{tabular}{|p{.1\linewidth}|p{.075\linewidth}|p{.075\linewidth}|p{.15\linewidth}|p{.2\linewidth}|p{.1\linewidth}|p{.1\linewidth}|}
\hline
\textbf{Model} & \textbf{1C} & \textbf{AC} & \textbf{Type of enhancement} & \textbf{Method} & \textbf{1C Improvement} & \textbf{AC Improvement} \\
\hline\hline

\multirow{9}{*}{Inception} & 24.37\% & 5.94\% & Baseline & Baseline & --- & --- \\
 & 31.80\% & 7.37\% & Resolution & Bicubic & 7.43\% & 1.43\% \\
 & 32.14\% & 7.52\% & Resolution & Bilinear & 7.77\% & 1.58\% \\
 & 26.10\% & 6.06\% & Deblurring & Blind Dec. & 1.73\% & 0.11\% \\
 & 25.32\% & 6.05\% & Deblurring & Deep Debl. & 0.95\% & 0.10\% \\
 & 33.33\% & 7.80\% & Deblurring & DDD & 8.96\% & 1.85\% \\
 & 31.53\% & 7.16\% & Resolution & Nearest neighbor & 7.16\% & 1.22\% \\
 & 32.90\% & 7.99\% & Resolution & SRCNN & 8.53\% & 2.04\% \\
 & 32.70\% & 7.07\% & Resolution & VDSR & 8.32\% & 1.13\% \\
\hline
\multirow{9}{*}{ResNet} & 30.60\% & 6.02\% & Baseline & Baseline & --- & --- \\
 & 30.86\% & 6.23\% & Resolution & Bicubic & 0.26\% & 0.21\% \\
 & 31.17\% & 6.05\% & Resolution & Bilinear & 0.58\% & 0.02\% \\
 & 24.53\% & 4.84\% & Deblurring & Blind Dec. & -6.07\% & -1.19\% \\
 & 32.77\% & 5.96\% & Deblurring & Deep Debl. & 2.18\% & -0.06\% \\
 & 31.66\% & 5.73\% & Deblurring & DDD & 1.06\% & -0.30\% \\
 & 30.40\% & 6.38\% & Resolution & Nearest neighbor & -0.20\% & 0.35\% \\
 & 30.98\% & 6.11\% & Resolution & SRCNN & 0.38\% & 0.08\% \\
 & 31.44\% & 5.34\% & Resolution & VDSR & 0.84\% & -0.68\% \\
\hline
\multirow{9}{*}{VGG16} & 28.96\% & 6.81\% & Baseline & Baseline & --- & --- \\
 & 28.94\% & 6.69\% & Resolution & Bicubic & -0.02\% & -0.12\% \\
 & 28.92\% & 6.50\% & Resolution & Bilinear & -0.04\% & -0.31\% \\
 & 25.37\% & 6.12\% & Deblurring & Blind Dec. & -3.59\% & -0.69\% \\
 & 30.53\% & 6.31\% & Deblurring & Deep Debl. & 1.57\% & -0.50\% \\
 & 30.56\% & 6.79\% & Deblurring & DDD & 1.60\% & -0.02\% \\
 & 28.67\% & 6.89\% & Resolution & Nearest neighbor & -0.30\% & 0.09\% \\
 & 32.67\% & 6.98\% & Resolution & SRCNN & 3.70\% & 0.18\% \\
 & 28.28\% & 6.06\% & Resolution & VDSR & -0.68\% & -0.74\% \\
\hline
\multirow{9}{*}{VGG19} & 29.68\% & 7.29\% & Baseline & Baseline & --- & --- \\
 & 29.56\% & 7.13\% & Resolution & Bicubic & -0.12\% & -0.15\% \\
 & 29.44\% & 6.93\% & Resolution & Bilinear & -0.24\% & -0.36\% \\
 & 24.57\% & 6.14\% & Deblurring & Blind Dec. & -5.11\% & -1.15\% \\
 & 32.08\% & 7.05\% & Deblurring & Deep Debl. & 2.40\% & -0.24\% \\
 & 31.43\% & 7.25\% & Deblurring & DDD & 1.75\% & -0.04\% \\
 & 29.32\% & 7.29\% & Resolution & Nearest neighbor & -0.36\% & 0.00\% \\
 & 32.51\% & 7.85\% & Resolution & SRCNN & 2.83\% & 0.56\% \\
 & 29.47\% & 6.70\% & Resolution & VDSR & -0.21\% & -0.59\% \\

\hline
\end{tabular}
\end{center}
\end{table}

\begin{table}[H]
\caption{Overall summary for the best and worst pre-processing algorithms for the UAV collection classification at rank 5\label{Supp-tab:YT_t5_overall_sum}}
\begin{center}
\begin{tabular}{|c|c|c|c|c|c|c|c|c|}
\hline
\textbf{Model} & \multicolumn{2}{|c|}{\textbf{Best 1C}} & \multicolumn{2}{|c|}{\textbf{Worst 1C}} & \multicolumn{2}{|c|}{\textbf{Best AC}} & \multicolumn{2}{|c|}{\textbf{Worst AC}} \\
\hline\hline

Inception & 8.96\% & DDD & 0.95\% & Deep Debl. & 2.04\% & SRCNN & 0.10\% & Deep Debl. \\
ResNet & 2.18\% & Deep Debl. & -6.07\% & Blind Dec. & 0.35\% & Nearest neighbor & -1.19\% & Blind Dec. \\
VGG16 & 3.70\% & SRCNN & -3.59\% & Blind Dec. & 0.18\% & SRCNN & -0.74\% & VDSR \\
VGG19 & 2.83\% & SRCNN & -5.11\% & Blind Dec. & 0.56\% & SRCNN & -1.15\% & Blind Dec. \\

\hline
\end{tabular}
\end{center}
\end{table}

~
\begin{table}[H]
\caption{Summary for the best and worst resolution enhancement algorithms for the UAV collection classification at rank 5\label{Supp-tab:YT_t5_overall_res}}
\begin{center}
\begin{tabular}{|c|c|c|c|c|c|c|c|c|}
\hline
\textbf{Model} & \multicolumn{2}{|c|}{\textbf{Best 1C}} & \multicolumn{2}{|c|}{\textbf{Worst 1C}} & \multicolumn{2}{|c|}{\textbf{Best AC}} & \multicolumn{2}{|c|}{\textbf{Worst AC}} \\
\hline\hline

Inception & 8.53\% & SRCNN & 7.16\% & Nearest neighbor & 2.04\% & SRCNN & 1.13\% & VDSR \\
ResNet & 0.84\% & VDSR & -0.20\% & Nearest neighbor & 0.35\% & Nearest neighbor & -0.68\% & VDSR \\
VGG16 & 3.70\% & SRCNN & -0.68\% & VDSR & 0.18\% & SRCNN & -0.74\% & VDSR \\
VGG19 & 2.83\% & SRCNN & -0.36\% & Nearest neighbor & 0.56\% & SRCNN & -0.59\% & VDSR \\

\hline
\end{tabular}
\end{center}
\end{table}

~
\begin{table}[H]
\caption{Overall summary for the best and worst deblurring algorithms for the UAV collection classification at rank 5\label{Supp-tab:YT_t5_overall_debl}}
\begin{center}
\begin{tabular}{|c|c|c|c|c|c|c|c|c|}
\hline
\textbf{Model} & \multicolumn{2}{|c|}{\textbf{Best 1C}} & \multicolumn{2}{|c|}{\textbf{Worst 1C}} & \multicolumn{2}{|c|}{\textbf{Best AC}} & \multicolumn{2}{|c|}{\textbf{Worst AC}} \\
\hline\hline

Inception & 8.96\% & DDD & 0.95\% & Deep Debl. & 1.85\% & DDD & 0.10\% & Deep Debl. \\
ResNet & 2.18\% & Deep Debl. & -6.07\% & Blind Dec. & -0.06\% & Deep Debl. & -1.19\% & Blind Dec. \\
VGG16 & 1.60\% & DDD & -3.59\% & Blind Dec. & -0.02\% & DDD & -0.69\% & Blind Dec. \\
VGG19 & 2.40\% & Deep Debl. & -5.11\% & Blind Dec. & -0.04\% & DDD & -1.15\% & Blind Dec. \\

\hline
\end{tabular}
\end{center}
\end{table}

\pagebreak
\subsection{Glider Collection}

~
\begin{table}[H]
 \caption{Details for the Glider Collection's classification results at rank 5} \label{Supp-tab:Kawa_t5_overall_dets}
\begin{center}
\begin{tabular}{|p{.1\linewidth}|p{.075\linewidth}|p{.075\linewidth}|p{.15\linewidth}|p{.2\linewidth}|p{.1\linewidth}|p{.1\linewidth}|}
\hline
\textbf{Model} & \textbf{1C} & \textbf{AC} & \textbf{Type of enhancement} & \textbf{Method} & \textbf{1C Improvement} & \textbf{AC Improvement} \\
\hline\hline

\multirow{9}{*}{Inception} & 35.07\% & 4.96\% & Baseline & Baseline & --- & ---\\
& 40.14\% & 3.75\% & Resolution & Bicubic & 5.07\% & -1.22\%\\
& 40.78\% & 3.95\% & Resolution & Bilinear & 5.71\% & -1.02\%\\
& 32.73\% & 3.25\% & Deblurring & Blind Dec. & -2.34\% & -1.72\%\\
& 35.10\% & 4.90\% & Deblurring & Deep Debl. & 0.03\% & -0.07\%\\
& 40.68\% & 4.24\% & Deblurring & DDD & 5.60\% & -0.72\%\\
& 39.65\% & 3.66\% & Resolution & Nearest neighbor & 4.58\% & -1.30\%\\
& 41.57\% & 5.39\% & Resolution & SRCNN & 6.50\% & 0.43\%\\
& 41.34\% & 4.09\% & Resolution & VDSR & 6.27\% & -0.88\%\\
\hline
\multirow{9}{*}{ResNet} & 39.38\% & 4.82\% & Baseline & Baseline & --- & ---\\
& 38.95\% & 4.43\% & Resolution & Bicubic & -0.43\% & -0.39\%\\
& 39.50\% & 4.28\% & Resolution & Bilinear & 0.12\% & -0.54\%\\
& 32.36\% & 3.00\% & Deblurring & Blind Dec. & -7.02\% & -1.82\%\\
& 41.40\% & 5.58\% & Deblurring & Deep Debl. & 2.02\% & 0.76\%\\
& 38.02\% & 4.33\% & Deblurring & DDD & -1.36\% & -0.49\%\\
& 38.37\% & 4.55\% & Resolution & Nearest neighbor & -1.01\% & -0.27\%\\
& 39.02\% & 4.84\% & Resolution & SRCNN & -0.36\% & 0.02\%\\
& 40.81\% & 3.42\% & Resolution & VDSR & 1.43\% & -1.40\%\\
\hline
\multirow{9}{*}{VGG16} & 36.09\% & 3.82\% & Baseline & Baseline & --- & ---\\
& 36.38\% & 4.01\% & Resolution & Bicubic & 0.29\% & 0.19\%\\
& 37.72\% & 4.02\% & Resolution & Bilinear & 1.64\% & 0.20\%\\
& 32.27\% & 2.84\% & Deblurring & Blind Dec. & -3.82\% & -0.98\%\\
& 38.30\% & 3.65\% & Deblurring & Deep Debl. & 2.21\% & -0.17\%\\
& 36.32\% & 3.47\% & Deblurring & DDD & 0.24\% & -0.34\%\\
& 35.45\% & 3.89\% & Resolution & Nearest neighbor & -0.63\% & 0.07\%\\
& 40.88\% & 4.33\% & Resolution & SRCNN & 4.79\% & 0.51\%\\
& 38.64\% & 4.54\% & Resolution & VDSR & 2.56\% & 0.73\%\\
\hline
\multirow{9}{*}{VGG19} & 35.22\% & 4.28\% & Baseline & Baseline & --- & ---\\
& 35.54\% & 4.39\% & Resolution & Bicubic & 0.32\% & 0.12\%\\
& 36.03\% & 4.42\% & Resolution & Bilinear & 0.80\% & 0.14\%\\
& 29.09\% & 3.90\% & Deblurring & Blind Dec. & -6.14\% & -0.37\%\\
& 37.00\% & 4.07\% & Deblurring & Deep Debl. & 1.78\% & -0.21\%\\
& 36.02\% & 4.01\% & Deblurring & DDD & 0.80\% & -0.27\%\\
& 34.91\% & 4.17\% & Resolution & Nearest neighbor & -0.32\% & -0.11\%\\
& 38.81\% & 4.75\% & Resolution & SRCNN & 3.58\% & 0.47\%\\
& 36.96\% & 4.32\% & Resolution & VDSR & 1.74\% & 0.04\% \\

\hline
\end{tabular}
\end{center}
\end{table}

~
\begin{table}[H]
\caption{Overall summary for the best and worst pre-processing algorithms for the Glider collection classification at rank 5\label{Supp-tab:Kawa_t5_overall_sum}}
\begin{center}
\begin{tabular}{|c|c|c|c|c|c|c|c|c|}
\hline
\textbf{Model} & \multicolumn{2}{|c|}{\textbf{Best 1C}} & \multicolumn{2}{|c|}{\textbf{Worst 1C}} & \multicolumn{2}{|c|}{\textbf{Best AC}} & \multicolumn{2}{|c|}{\textbf{Worst AC}} \\
\hline\hline

Inception & 6.50\% & SRCNN & -2.34\% & Blind Dec. & 0.43\% & SRCNN & -1.72\% & Blind Dec. \\
ResNet & 2.02\% & Deep Debl. & -7.02\% & Blind Dec. & 0.76\% & Deep Debl. & -1.82\% & Blind Dec. \\
VGG16 & 4.79\% & SRCNN & -3.82\% & Blind Dec. & 0.73\% & VDSR & -0.98\% & Blind Dec. \\
VGG19 & 3.58\% & SRCNN & -6.14\% & Blind Dec. & 0.47\% & SRCNN & -0.37\% & Blind Dec. \\

\hline
\end{tabular}
\end{center}
\end{table}

~
\begin{table}[H]
\caption{Summary of the best and worst resolution enhancement algorithms for the Glider collection classification at rank 5\label{Supp-tab:Kawa_t5_overall_res}}
\begin{center}
\begin{tabular}{|c|c|c|c|c|c|c|c|c|}
\hline
\textbf{Model} & \multicolumn{2}{|c|}{\textbf{Best 1C}} & \multicolumn{2}{|c|}{\textbf{Worst 1C}} & \multicolumn{2}{|c|}{\textbf{Best AC}} & \multicolumn{2}{|c|}{\textbf{Worst AC}} \\
\hline\hline

Inception & 6.50\% & SRCNN & 4.58\% & Nearest neighbor & 0.43\% & SRCNN & -1.30\% & Nearest neighbor \\
ResNet & 1.43\% & VDSR & -1.01\% & Nearest neighbor & 0.02\% & SRCNN & -1.40\% & VDSR \\
VGG16 & 4.79\% & SRCNN & -0.63\% & Nearest neighbor & 0.73\% & VDSR & 0.07\% & Nearest neighbor \\
VGG19 & 3.58\% & SRCNN & -0.32\% & Nearest neighbor & 0.47\% & SRCNN & -0.11\% & Nearest neighbor \\

\hline
\end{tabular}
\end{center}
\end{table}

~
\begin{table}[H]
\caption{Overall summary for the best and worst deblurring algorithms for the Glider collection classification at rank 5\label{Supp-tab:Kawa_t5_overall_debl}}
\begin{center}
\begin{tabular}{|c|c|c|c|c|c|c|c|c|}
\hline
\textbf{Model} & \multicolumn{2}{|c|}{\textbf{Best 1C}} & \multicolumn{2}{|c|}{\textbf{Worst 1C}} & \multicolumn{2}{|c|}{\textbf{Best AC}} & \multicolumn{2}{|c|}{\textbf{Worst AC}} \\
\hline\hline

Inception & 5.60\% & DDD & -2.34\% & Blind Dec. & -0.07\% & Deep Debl. & -1.72\% & Blind Dec. \\
ResNet & 2.02\% & Deep Debl. & -7.02\% & Blind Dec. & 0.76\% & Deep Debl. & -1.82\% & Blind Dec. \\
VGG16 & 2.21\% & Deep Debl. & -3.82\% & Blind Dec. & -0.17\% & Deep Debl. & -0.98\% & Blind Dec. \\
VGG19 & 1.78\% & Deep Debl. & -6.14\% & Blind Dec. & -0.21\% & Deep Debl. & -0.37\% & Blind Dec. \\
\hline
\end{tabular}
\end{center}
\end{table}

\pagebreak
\subsection{Ground Collection}
~
\begin{table}[H]
 \caption{Details for the Ground Collection's classification results at rank 5} \label{Supp-tab:Ground_t5_overall_dets}
\begin{center}
\begin{tabular}{|p{.1\linewidth}|p{.075\linewidth}|p{.075\linewidth}|p{.15\linewidth}|p{.2\linewidth}|p{.1\linewidth}|p{.1\linewidth}|}
\hline
\textbf{Model} & \textbf{1C} & \textbf{AC} & \textbf{Type of enhancement} & \textbf{Method} & \textbf{1C Improvement} & \textbf{AC Improvement} \\
\hline\hline

\multirow{9}{*}{Inception} & 54.86\% & 49.16\% & Baseline & Baseline & --- & --- \\
 & 54.26\% & 45.80\% & Resolution & Bicubic & -0.60\% & -3.36\% \\
 & 53.78\% & 45.33\% & Resolution & Bilinear & -1.08\% & -3.84\% \\
 & 45.16\% & 37.08\% & Deblurring & Blind Dec. & -9.70\% & -12.09\% \\
 & 53.98\% & 45.93\% & Deblurring & Deep Debl. & -0.88\% & -3.24\% \\
 & 52.98\% & 44.71\% & Deblurring & DDD & -1.88\% & -4.46\% \\
 & 54.26\% & 45.80\% & Resolution & Nearest neighbor & -0.60\% & -3.36\% \\
 & 53.43\% & 44.60\% & Resolution & SRCNN & -1.43\% & -4.56\% \\
 & 55.67\% & 46.78\% & Resolution & VDSR & 0.81\% & -2.39\% \\
\hline
\multirow{9}{*}{ResNet} & 56.48\% & 46.39\% & Baseline & Baseline & --- & --- \\
 & 56.93\% & 47.47\% & Resolution & Bicubic & 0.45\% & 1.08\% \\
 & 57.23\% & 47.47\% & Resolution & Bilinear & 0.75\% & 1.08\% \\
 & 42.03\% & 35.01\% & Deblurring & Blind Dec. & -14.46\% & -11.38\% \\
 & 55.23\% & 45.19\% & Deblurring & Deep Debl. & -1.26\% & -1.20\% \\
 & 55.05\% & 44.71\% & Deblurring & DDD & -1.43\% & -1.67\% \\
 & 56.93\% & 47.47\% & Resolution & Nearest neighbor & 0.45\% & 1.08\% \\
 & 54.09\% & 44.15\% & Resolution & SRCNN & -2.39\% & -2.23\% \\
 & 56.15\% & 46.71\% & Resolution & VDSR & -0.33\% & 0.33\% \\
\hline
\multirow{9}{*}{VGG16} & 57.14\% & 52.89\% & Baseline & Baseline & --- & --- \\
 & 56.79\% & 52.79\% & Resolution & Bicubic & -0.34\% & -0.11\% \\
 & 57.02\% & 51.74\% & Resolution & Bilinear & -0.12\% & -1.15\% \\
 & 42.47\% & 37.83\% & Deblurring & Blind Dec. & -14.66\% & -15.06\% \\
 & 57.25\% & 52.88\% & Deblurring & Deep Debl. & 0.12\% & -0.01\% \\
 & 57.26\% & 53.32\% & Deblurring & DDD & 0.12\% & 0.43\% \\
 & 56.79\% & 52.79\% & Resolution & Nearest neighbor & -0.34\% & -0.11\% \\
 & 54.29\% & 49.08\% & Resolution & SRCNN & -2.84\% & -3.81\% \\
 & 58.64\% & 54.63\% & Resolution & VDSR & 1.51\% & 1.74\% \\
\hline
\multirow{9}{*}{VGG19} & 53.37\% & 48.34\% & Baseline & Baseline & --- & --- \\
 & 53.23\% & 48.20\% & Resolution & Bicubic & -0.14\% & -0.14\% \\
 & 53.60\% & 48.28\% & Resolution & Bilinear & 0.23\% & -0.07\% \\
 & 41.78\% & 36.89\% & Deblurring & Blind Dec. & -11.59\% & -11.45\% \\
 & 53.53\% & 48.50\% & Deblurring & Deep Debl. & 0.16\% & 0.16\% \\
 & 55.21\% & 50.00\% & Deblurring & DDD & 1.84\% & 1.66\% \\
 & 53.23\% & 48.20\% & Resolution & Nearest neighbor & -0.14\% & -0.14\% \\
 & 54.26\% & 50.67\% & Resolution & SRCNN & 0.89\% & 2.33\% \\
 & 56.93\% & 51.15\% & Resolution & VDSR & 3.56\% & 2.80\% \\

\hline
\end{tabular}
\end{center}
\end{table}

~
\begin{table}[H]
\caption{Overall summary of the best and worst pre-processing algorithms for the Ground collection classification at rank 5\label{Supp-tab:Ground_t5_overall_sum}}
\begin{center}
\begin{tabular}{|c|c|c|c|c|c|c|c|c|}
\hline
\textbf{Model} & \multicolumn{2}{|c|}{\textbf{Best 1C}} & \multicolumn{2}{|c|}{\textbf{Worst 1C}} & \multicolumn{2}{|c|}{\textbf{Best AC}} & \multicolumn{2}{|c|}{\textbf{Worst AC}} \\
\hline\hline

Inception & 0.81\% & VDSR & -9.70\% & Blind Dec. & -2.39\% & VDSR & -12.09\% & Blind Dec.\\
ResNet & 0.75\% & Bilinear & -14.46\% & Blind Dec. & 1.08\% & Bilinear & -11.38\% & Blind Dec.\\
VGG16 & 1.51\% & VDSR & -14.66\% & Blind Dec. & 1.74\% & VDSR & -15.06\% & Blind Dec.\\
VGG19 & 3.56\% & VDSR & -11.59\% & Blind Dec. & 2.80\% & VDSR & -11.45\% & Blind Dec.\\

\hline
\end{tabular}
\end{center}
\end{table}

~
\begin{table}[H]
\caption{Summary of the best \& worst resolution enhancement algorithms for the Ground collection classification at rank 5\label{Supp-tab:Ground_t5_overall_res}}
\begin{center}
\begin{tabular}{|c|c|c|c|c|c|c|c|c|}
\hline
\textbf{Model} & \multicolumn{2}{|c|}{\textbf{Best 1C}} & \multicolumn{2}{|c|}{\textbf{Worst 1C}} & \multicolumn{2}{|c|}{\textbf{Best AC}} & \multicolumn{2}{|c|}{\textbf{Worst AC}} \\
\hline\hline

Inception & 0.81\% & VDSR & -1.43\% & SRCNN & -2.39\% & VDSR & -4.56\% & SRCNN \\
ResNet & 0.75\% & Bilinear & -2.39\% & SRCNN & 1.08\% & Bilinear & -2.23\% & SRCNN \\
VGG16 & 1.51\% & VDSR & -2.84\% & SRCNN & 1.74\% & VDSR & -3.81\% & SRCNN \\
VGG19 & 3.56\% & VDSR & -0.14\% & Bicubic & 2.80\% & VDSR & -0.14\% & Bicubic \\

\hline
\end{tabular}
\end{center}
\end{table}

~
\begin{table}[H]
\caption{Overall summary for the best and worst deblurring algorithms for the Ground collection classification at rank 5\label{Supp-tab:Ground_t5_overall_debl}}
\begin{center}
\begin{tabular}{|c|c|c|c|c|c|c|c|c|}
\hline
\textbf{Model} & \multicolumn{2}{|c|}{\textbf{Best 1C}} & \multicolumn{2}{|c|}{\textbf{Worst 1C}} & \multicolumn{2}{|c|}{\textbf{Best AC}} & \multicolumn{2}{|c|}{\textbf{Worst AC}} \\
\hline\hline

Inception & -0.88\% & Deep Debl. & -9.70\% & Blind Dec. & -3.24\% & Deep Debl. & -12.09\% & Blind Dec. \\
ResNet & -1.26\% & Deep Debl. & -14.46\% & Blind Dec. & -1.20\% & Deep Debl. & -11.38\% & Blind Dec. \\
VGG16 & 0.12\% & DDD & -14.66\% & Blind Dec. & 0.43\% & DDD & -15.06\% & Blind Dec. \\
VGG19 & 1.84\% & DDD & -11.59\% & Blind Dec. & 1.66\% & DDD & -11.45\% & Blind Dec. \\

\hline
\end{tabular}
\end{center}
\end{table}

\pagebreak
\subsubsection{Impact of weather conditions}

~
\begin{table}[H]
 \caption{Details for the Ground Collection's classification results (under different weather conditions) at rank 5} \label{Supp-tab:Ground_t5_weather_overall_dets}
\begin{center}
\begin{tabular}{|p{.1\linewidth}|p{.075\linewidth}|p{.075\linewidth}|p{.15\linewidth}|p{.2\linewidth}|p{.1\linewidth}|p{.1\linewidth}|}
\hline
\textbf{Model} & \textbf{1C} & \textbf{AC} & \textbf{Type of enhancement} & \textbf{Method} & \textbf{1C Improvement} & \textbf{AC Improvement} \\
\hline\hline

\multirow{9}{*}{Cloudy} & 69.88\% & 65.17\% & Baseline & Baseline & --- & --- \\
 & 65.69\% & 61.33\% & Resolution & Bicubic & -4.19\% & -3.85\% \\
 & 66.73\% & 61.35\% & Resolution & Bilinear & -3.15\% & -3.82\% \\
 & 47.13\% & 42.81\% & Deblurring & Blind Dec. & -22.75\% & -22.36\% \\
 & 65.93\% & 61.45\% & Deblurring & Deep Debl. & -3.95\% & -3.72\% \\
 & 65.41\% & 60.40\% & Deblurring & DDD & -4.47\% & -4.77\% \\
 & 65.69\% & 61.33\% & Resolution & Nearest neighbor & -4.19\% & -3.85\% \\
 & 63.64\% & 57.63\% & Resolution & SRCNN & -6.24\% & -7.54\% \\
 & 65.79\% & 61.85\% & Resolution & VDSR & -4.09\% & -3.32\% \\
\hline
\multirow{9}{*}{Rainy} & 0.06\% & 0.06\% & Baseline & Baseline & --- & --- \\
 & 0.00\% & 0.00\% & Resolution & Bicubic & -0.06\% & -0.06\% \\
 & 0.06\% & 0.06\% & Resolution & Bilinear & 0.00\% & 0.00\% \\
 & 1.35\% & 1.35\% & Deblurring & Blind Dec. & 1.28\% & 1.28\% \\
 & 0.00\% & 0.00\% & Deblurring & Deep Debl. & -0.06\% & -0.06\% \\
 & 0.00\% & 0.00\% & Deblurring & DDD & -0.06\% & -0.06\% \\
 & 0.00\% & 0.00\% & Resolution & Nearest neighbor & -0.06\% & -0.06\% \\
 & 0.32\% & 0.32\% & Resolution & SRCNN & 0.26\% & 0.26\% \\
 & 0.00\% & 0.00\% & Resolution & VDSR & -0.06\% & -0.06\% \\
\hline
\multirow{9}{*}{Snow} & 40.51\% & 40.51\% & Baseline & Baseline & --- & --- \\
 & 42.68\% & 42.68\% & Resolution & Bicubic & 2.17\% & 2.17\% \\
 & 42.61\% & 42.61\% & Resolution & Bilinear & 2.10\% & 2.10\% \\
 & 32.13\% & 32.13\% & Deblurring & Blind Dec. & -8.38\% & -8.38\% \\
 & 42.55\% & 42.55\% & Deblurring & Deep Debl. & 2.03\% & 2.03\% \\
 & 42.31\% & 42.31\% & Deblurring & DDD & 1.79\% & 1.79\% \\
 & 42.68\% & 42.68\% & Resolution & Nearest neighbor & 2.17\% & 2.17\% \\
 & 42.44\% & 42.44\% & Resolution & SRCNN & 1.93\% & 1.93\% \\
 & 42.62\% & 42.62\% & Resolution & VDSR & 2.10\% & 2.10\% \\
\hline
\multirow{9}{*}{Sunny} & 52.73\% & 44.83\% & Baseline & Baseline & --- & --- \\
 & 53.73\% & 44.96\% & Resolution & Bicubic & 1.00\% & 0.14\% \\
 & 53.51\% & 44.40\% & Resolution & Bilinear & 0.78\% & -0.42\% \\
 & 43.22\% & 35.34\% & Deblurring & Blind Dec. & -9.51\% & -9.49\% \\
 & 53.19\% & 44.26\% & Deblurring & Deep Debl. & 0.46\% & -0.57\% \\
 & 53.62\% & 44.79\% & Deblurring & DDD & 0.89\% & -0.04\% \\
 & 53.73\% & 44.96\% & Resolution & Nearest neighbor & 1.00\% & 0.14\% \\
 & 52.56\% & 44.18\% & Resolution & SRCNN & -0.17\% & -0.65\% \\
 & 56.09\% & 46.71\% & Resolution & VDSR & 3.36\% & 1.89\% \\

\hline
\end{tabular}
\end{center}
\end{table}

~
\begin{table}[H]
\caption{Overall summary for the best and worst pre-processing algorithms to deal with adverse weather conditions for the Ground collection classification at rank 5\label{Supp-tab:Ground_t5_weath_overall_sum}}
\begin{center}
\begin{tabular}{|c|c|c|c|c|c|c|c|c|}
\hline
\textbf{Weather} & \multicolumn{2}{|c|}{\textbf{Best 1C}} & \multicolumn{2}{|c|}{\textbf{Worst 1C}} & \multicolumn{2}{|c|}{\textbf{Best AC}} & \multicolumn{2}{|c|}{\textbf{Worst AC}} \\
\hline\hline

Cloudy & -3.15\% & Bilinear & -22.75\% & Blind Dec. & -3.32\% & VDSR & -22.36\% & Blind Dec. \\
Rainy & 1.28\% & Blind Dec. & -0.06\% & Bicubic & 1.28\% & Blind Dec. & -0.06\% & Bicubic \\
Snow & 2.17\% & Bicubic & -8.38\% & Blind Dec. & 2.17\% & Bicubic & -8.38\% & Blind Dec. \\
Sunny & 3.36\% & VDSR & -9.51\% & Blind Dec. & 1.89\% & VDSR & -9.49\% & Blind Dec. \\

\hline
\end{tabular}
\end{center}
\end{table}

~
\begin{table}[H]
\caption{Summary for the best and worst resolution enhancement algorithms to deal with adverse weather conditions for the Ground collection classification at rank 5\label{Supp-tab:Ground_t5_weath_overall_res}}
\begin{center}
\begin{tabular}{|c|c|c|c|c|c|c|c|c|}
\hline
\textbf{Weather} & \multicolumn{2}{|c|}{\textbf{Best 1C}} & \multicolumn{2}{|c|}{\textbf{Worst 1C}} & \multicolumn{2}{|c|}{\textbf{Best AC}} & \multicolumn{2}{|c|}{\textbf{Worst AC}} \\
\hline\hline

Cloudy & -3.15\% & Bilinear & -6.24\% & SRCNN & -3.32\% & VDSR & -7.54\% & SRCNN \\
Rainy & 0.26\% & SRCNN & -0.06\% & Bicubic & 0.26\% & SRCNN & -0.06\% & Bicubic \\
Snow & 2.17\% & Bicubic & 1.93\% & SRCNN & 2.17\% & Bicubic & 1.93\% & SRCNN \\
Sunny & 3.36\% & VDSR & -0.17\% & SRCNN & 1.89\% & VDSR & -0.65\% & SRCNN \\

\hline
\end{tabular}
\end{center}
\end{table}

~
\begin{table}[H]
\caption{Overall summary for the best and worst deblurring algorithms to deal with adverse weather conditions for the Ground collection classification at rank 5\label{Supp-tab:Ground_t5_weath_overall_debl}}
\begin{center}
\begin{tabular}{|c|c|c|c|c|c|c|c|c|}
\hline
\textbf{Weather} & \multicolumn{2}{|c|}{\textbf{Best 1C}} & \multicolumn{2}{|c|}{\textbf{Worst 1C}} & \multicolumn{2}{|c|}{\textbf{Best AC}} & \multicolumn{2}{|c|}{\textbf{Worst AC}} \\
\hline\hline

Cloudy & -3.95\% & Deep Debl. & -22.75\% & Blind Dec. & -3.72\% & Deep Debl. & -22.36\% & Blind Dec. \\
Rainy & 1.28\% & Blind Dec. & -0.06\% & Bicubic & 1.28\% & Blind Dec. & -0.06\% & Bicubic \\
Snow & 2.03\% & Deep Debl. & -8.38\% & Blind Dec. & 2.03\% & Deep Debl. & -8.38\% & Blind Dec. \\
Sunny & 0.89\% & DDD & -9.51\% & Blind Dec. & -0.04\% & DDD & -9.49\% & Blind Dec. \\

\hline
\end{tabular}
\end{center}
\end{table}

\pagebreak
\section{Rank 1 Classification results}

\begin{figure}[ht]
\begin{center}
\pgfplotstableread[row sep=\\,col sep=&]{
    Method & Baseline & Bilinear & Bicubic & NearestNeighbor & VDSR & SRCNN & DeepDeblurring & BlindDeconvolution & DynamicDeepDeblurring \\
    Inception & 11.5 & 15.3 & 14.8 & 14.6 & 14.3 & 15.6 & 11.7 & 11.7 & 15.3 \\
	ResNet & 14.7 & 14.3 & 14.4 & 14.3 & 13.1 & 15.0 & 15.0 & 10.6 & 14.9 \\
	VGG16 & 9.8 & 9.8 & 9.7 & 9.8 & 9.6 & 11.9 & 10.1 & 9.2 & 10.0 \\
	VGG19 & 11.1 & 11.4 & 11.1 & 11.0 & 11.2 & 12.6 & 11.8 & 9.3 & 11.4 \\
    }\YTTOneData

\begin{tikzpicture}
    \begin{axis}[
            title={UAV Collection: Rank 1 Classification},
            ylabel={Correct classification rate [\%]},
            xlabel={Pre-trained network},
            ybar,
            bar width=.3cm,
            width=\textwidth,
            height=.45\textwidth,
            enlarge x limits=0.16,
            minor y tick num= 1,
            minor grid style=loosely dotted,
            yminorgrids = true, 
            legend style={at={(0.5,-0.19)},
                anchor=north,legend columns=5},
            legend image code/.code={%
              \draw[#1] (0cm,-0.1cm) rectangle (0.6cm,0.1cm);
            }, 
            symbolic x coords={Inception,ResNet,VGG16,VGG19},
            xtick=data,
            ymin=0,ymax=20,
        ]
        \addplot [color=black,fill=red] table[x=Method,y=Baseline]{\YTTOneData};
        \addplot [color=black, fill=blue] table[x=Method,y=Bilinear]{\YTTOneData};
        \addplot [color=black, fill=green!70!black] table[x=Method,y=Bicubic]{\YTTOneData};
        \addplot [color=black, fill=pink] table[x=Method,y=NearestNeighbor]{\YTTOneData};
        \addplot [color=black, fill=magenta] table[x=Method,y=VDSR]{\YTTOneData};
        \addplot [color=black, fill=orange] table[x=Method,y=SRCNN]{\YTTOneData};
        \addplot [color=black, fill=purple] table[x=Method,y=BlindDeconvolution]{\YTTOneData};
        \addplot [color=black, fill=cyan] table[x=Method,y=DeepDeblurring]{\YTTOneData};
        \addplot [color=black, fill=yellow!90!black] table[x=Method,y=DynamicDeepDeblurring]{\YTTOneData};

        \legend{Baseline, Bilinear, Bicubic, NearestNeighbor, VDSR, SRCNN, BlindDeconvolution, DeepDeblurring, DynamicDeepDeblurring}
    \end{axis}
\end{tikzpicture}
\end{center}
\vspace{-5mm}
   \caption{Comparison of classification rates at rank 1 for the UAV Collection after applying several resolution enhancement and deblurring techniques.}
\label{Supp-fig:graph:YT_T1}
\end{figure}

\begin{figure}[ht]
\begin{center}
\pgfplotstableread[row sep=\\,col sep=&]{
    Method & Baseline & Bilinear & Bicubic & NearestNeighbor & VDSR & SRCNN & DeepDeblurring & BlindDeconvolution & DynamicDeepDeblurring \\
    Inception & 20.5 & 21.7 & 21.4 & 21.1 & 22.5 & 23.7 & 21.2 & 17.7 & 22.6 \\
    ResNet & 19.4 & 19.8 & 19.5 & 19.4 & 19.9 & 19.8 & 20.7 & 14.0 & 19.4 \\
    VGG16 & 19.2 & 19.6 & 19.4 & 19.1 & 20.6 & 20.5 & 19.6 & 15.1 & 19.1 \\
    VGG19 & 19.2 & 19.8 & 19.8 & 19.4 & 20.1 & 19.4 & 19.4 & 12.9 & 18.9 \\
    }\KTOneData

\begin{tikzpicture}
    \begin{axis}[
            title={Glider Collection: Rank 1 Classification},
            ylabel={Correct classification rate [\%]},
            xlabel={Pre-trained network},
            ybar,
            bar width=.3cm,
            width=\textwidth,
            height=.45\textwidth,
            enlarge x limits=0.16,
            minor y tick num= 1,
            minor grid style=loosely dotted,
            yminorgrids = true, 
            legend style={at={(0.5,-0.19)},
                anchor=north,legend columns=5},
            legend image code/.code={%
              \draw[#1] (0cm,-0.1cm) rectangle (0.6cm,0.1cm);
            }, 
            symbolic x coords={Inception,ResNet,VGG16,VGG19},
            xtick=data,
            ymin=0,ymax=30,
        ]
        \addplot [color=black,fill=red] table[x=Method,y=Baseline]{\KTOneData};
        \addplot [color=black, fill=blue] table[x=Method,y=Bilinear]{\KTOneData};
        \addplot [color=black, fill=green!70!black] table[x=Method,y=Bicubic]{\KTOneData};
        \addplot [color=black, fill=pink] table[x=Method,y=NearestNeighbor]{\KTOneData};
        \addplot [color=black, fill=magenta] table[x=Method,y=VDSR]{\KTOneData};
        \addplot [color=black, fill=orange] table[x=Method,y=SRCNN]{\KTOneData};
        \addplot [color=black, fill=purple] table[x=Method,y=BlindDeconvolution]{\KTOneData};
        \addplot [color=black, fill=cyan] table[x=Method,y=DeepDeblurring]{\KTOneData};
        \addplot [color=black, fill=yellow!90!black] table[x=Method,y=DynamicDeepDeblurring]{\KTOneData};

        \legend{Baseline, Bilinear, Bicubic, NearestNeighbor, VDSR, SRCNN, BlindDeconvolution, DeepDeblurring, DynamicDeepDeblurring}
    \end{axis}
\end{tikzpicture}
\end{center}
\vspace{-5mm}
   \caption{Comparison of classification rates at rank 1 for the Glider Collection after applying several resolution enhancement and deblurring techniques.}
\label{Supp-fig:graph:K_T1}
\end{figure}

\begin{figure}[ht]
\begin{center}
\pgfplotstableread[row sep=\\,col sep=&]{
    Method & Baseline & Bilinear & Bicubic & NearestNeighbor & VDSR & SRCNN & DeepDeblurring & BlindDeconvolution & DynamicDeepDeblurring \\
    Inception & 34.4 & 36.0 & 35.9 & 35.9 & 36.1 & 36.3 & 33.6 & 29.3 & 35.0 \\
    ResNet & 34.9 & 36.8 & 36.2 & 36.2 & 37.2 & 33.8 & 36.1 & 26.5 & 35.6 \\
    VGG16 & 28.8 & 28.5 & 28.9 & 28.9 & 30.2 & 29.1 & 29.7 & 22.6 & 28.8 \\
    VGG19 & 30.8 & 30.0 & 30.5 & 30.5 & 32.3 & 31.0 & 31.5 & 25.2 & 30.3 \\
    }\GTOneData

\begin{tikzpicture}
    \begin{axis}[
            title={Ground Collection: Rank 1 Classification},
            ylabel={Correct classification rate [\%]},
            xlabel={Pre-trained network},
            ybar,
            bar width=.3cm,
            width=\textwidth,
            height=.45\textwidth,
            enlarge x limits=0.16,
            minor y tick num= 1,
            minor grid style=loosely dotted,
            yminorgrids = true, 
            legend style={at={(0.5,-0.19)},
                anchor=north,legend columns=5},
            legend image code/.code={%
              \draw[#1] (0cm,-0.1cm) rectangle (0.6cm,0.1cm);
            }, 
            symbolic x coords={Inception,ResNet,VGG16,VGG19},
            xtick=data,
            ymin=0,ymax=50,
        ]
        \addplot [color=black,fill=red] table[x=Method,y=Baseline]{\GTOneData};
        \addplot [color=black, fill=blue] table[x=Method,y=Bilinear]{\GTOneData};
        \addplot [color=black, fill=green!70!black] table[x=Method,y=Bicubic]{\GTOneData};
        \addplot [color=black, fill=pink] table[x=Method,y=NearestNeighbor]{\GTOneData};
        \addplot [color=black, fill=magenta] table[x=Method,y=VDSR]{\GTOneData};
        \addplot [color=black, fill=orange] table[x=Method,y=SRCNN]{\GTOneData};
        \addplot [color=black, fill=purple] table[x=Method,y=BlindDeconvolution]{\GTOneData};
        \addplot [color=black, fill=cyan] table[x=Method,y=DeepDeblurring]{\GTOneData};
        \addplot [color=black, fill=yellow!90!black] table[x=Method,y=DynamicDeepDeblurring]{\GTOneData};

        \legend{Baseline, Bilinear, Bicubic, NearestNeighbor, VDSR, SRCNN, BlindDeconvolution, DeepDeblurring, DynamicDeepDeblurring}
    \end{axis}
\end{tikzpicture}
\end{center}
\vspace{-5mm}
   \caption{Comparison of classification rates at rank 1 for the Ground Collection after applying several resolution enhancement and deblurring techniques.}
\label{Supp-fig:graph:G_T1}
\end{figure}

\section{Sample Videos}
We include three sample videos from UG$^2$, one per collection. Given the size restrictions, the videos included are segments from the original videos, and the video quality and speed (in the case of the UAV Collection sample) were modified to reduce the file size.